\documentclass[final,12pt]{elsarticle}

\usepackage{booktabs, tabularx, caption}
\usepackage{amssymb}
\usepackage{amsmath}
\usepackage{bm}
\usepackage{xcolor}

\usepackage{tabularx}
\usepackage{float}

\usepackage{url}
\usepackage{adjustbox}
\usepackage[colorlinks,
            linkcolor=blue,
            anchorcolor=blue,
            citecolor=blue]{hyperref}

\journal{Nuclear Physics B}

\begin{document}
\begin{frontmatter}



\title{Towards Practical Alzheimer’s Disease Diagnosis:\\ A Lightweight and Interpretable Spiking Neural Model} 


\author[hdu]{Changwei Wu\fnref{equal1}}
\author[thu]{Yifei Chen\fnref{equal1}\corref{cor1}}
\author[hdu]{Yuxin Du}
\author[hdu]{Jinying Zong}
\author[hdu]{Jie Dong}
\author[thu]{Mingxuan Liu}
\author[hdu]{Feiwei Qin\corref{cor1}}
\author[hdu]{Yong Peng}
\author[hdu]{Jin Fan}
\author[wcm]{Chaomiao Wang}

\affiliation[hdu]{organization={Hangzhou Dianzi University},
    city={Hangzhou},
    country={China}}

\affiliation[thu]{organization={Tsinghua University},
            city={Beijing},
            country={China}}

\affiliation[wcm]{organization={Shenzhen Research Institute of Big Data},
            city={Shenzhen},
            country={China}}

\fntext[equal1]{These authors contributed equally to this work.}
\cortext[cor1]{Corresponding authors: Yifei Chen (justlfc03@gmail.com), Feiwei Qin (qinfeiwei@hdu.edu.cn).}

\begin{abstract}
Early diagnosis of Alzheimer’s Disease (AD), particularly at the mild cognitive impairment stage, is essential for timely intervention. However, this process faces significant barriers, including reliance on subjective assessments and the high cost of advanced imaging techniques. While deep learning offers automated solutions to improve diagnostic accuracy, its widespread adoption remains constrained due to high energy requirements and computational demands, particularly in resource-limited settings. Spiking neural networks (SNNs) provide a promising alternative, as their brain-inspired design is well-suited to model the sparse and event-driven patterns characteristic of neural degeneration in AD. These networks offer the potential for developing interpretable, energy-efficient diagnostic tools. Despite their advantages, existing SNNs often suffer from limited expressiveness and challenges in stable training, which reduce their effectiveness in handling complex medical tasks. To address these shortcomings, we introduce FasterSNN, a hybrid neural architecture that combines biologically inspired Leaky Integrate-and-Fire (LIF) neurons with region-adaptive convolution and multi-scale spiking attention mechanisms. This approach facilitates efficient, sparse processing of 3D MRI data while maintaining high diagnostic accuracy. Experimental results on benchmark datasets reveal that FasterSNN delivers competitive performance with significantly enhanced efficiency and training stability, highlighting its potential for practical application in AD screening. Our source code is available at \href{https://github.com/wuchangw/FasterSNN}{https://github.com/wuchangw/FasterSNN}.
\end{abstract}




\end{frontmatter}



\section{Introduction}
\label{sec1}
Alzheimer's Disease (AD) is a progressive and irreversible neurodegenerative disorder characterized by cognitive decline, behavioral changes, and psychological symptoms such as memory loss, anxiety, depression, and hallucinations \cite{40}. As the disease progresses, these symptoms worsen, with severe cases often leading to complete loss of independence and, in some instances, mortality\cite{oh2025cerebrospinal}. The global aging population has further exacerbated the prevalence of AD, with the number of affected individuals expected to rise from 55 million in 2019 to 139 million by 2050. This growing burden places significant financial and emotional strain on families while also overwhelming healthcare systems and senior care services, creating a serious public health challenge worldwide \cite{38}. Typically, AD begins as mild cognitive impairment (MCI) \cite{39}, an early stage marked by subtle cognitive deficits that gradually progress to severe dementia. In its late stages, patients lose the ability to perform daily tasks and require extensive care. However, early detection and treatment during the MCI stage can help slow the progression of the disease and, in some cases, partially restore cognitive function\cite{zhu2025towards}. Therefore, accurate diagnosis at the MCI stage is critically important for effective disease management and improving patient outcomes.

Although early diagnosis is crucial for managing AD, particularly at the MCI stage, accurate identification remains a significant challenge in clinical practice\cite{kang20233dnest}. Widely used tools such as the AD Assessment Scale-Cognitive Subscale (ADAS-Cog) can effectively evaluate cognitive domains like memory, language, and executive function. However, with only 12 test items, ADAS-Cog lacks sensitivity, relies heavily on subjective judgment, and struggles to distinguish the gradual progression from MCI to AD with precision \cite{1}. While advanced imaging techniques such as structural magnetic resonance imaging (MRI), positron emission tomography/computed tomography (PET-CT), and PET/MRI can accurately detect brain atrophy and molecular pathological changes \cite{37}, their high costs and reliance on specialized expertise limit their accessibility and widespread use in clinical settings\cite{chen2022neuroimaging}. Additionally, the clinical symptoms of non-AD dementias,  such as Lewy Body Dementia and Frontotemporal Dementia, often overlap significantly with those of AD, leading to alarmingly high misdiagnosis rates \cite{2}. These challenges not only increase the diagnostic burden on physicians but also delay patients’ access to accurate diagnoses and timely interventions, further complicating disease management.

Despite PET demonstrating higher specificity in real-world diagnostic applications, its high costs place a significant financial burden on patients and their families. Moreover, PET imaging requires skilled professionals, further limiting its accessibility. In comparison, MRI is emerging as a more practical alternative due to its lower cost and consistent imaging quality \cite{zhang2024tc,ge2024tc,chen2024scunet++}. While cognitive assessment scales are affordable, functional, and easy to use, their reliability can be affected by factors such as a patient’s educational background, psychological state, and regional disparities, particularly in areas with high illiteracy rates. This highlights the need for diagnostic methods that combine economic feasibility with objective accuracy. MRI, as a straightforward imaging modality, offers clear advantages in terms of cost and equipment maintenance while preserving the objectivity of medical imaging. In this study, we focus exclusively on MRI unimodal data to avoid the subjective biases and resource inefficiencies associated with multimodal fusion. Our goal is to develop a diagnostic model that is not only more accurate and cost-effective but also practical and scalable for widespread deployment.

In recent years, artificial intelligence (AI) technologies, particularly deep learning methods, have demonstrated significant potential in assisting with AD diagnosis \cite{41,43,44}. These methods enable efficient and objective identification of critical imaging features, such as hippocampal atrophy and frontal lobe abnormalities \cite{42}. Despite their promise, artificial neural network (ANN) models often require substantial computational resources, high memory footprint and latency and consume significant power, making them impractical for widespread use in primary healthcare settings and resource-limited regions. Bio-inspired spiking neural networks (SNNs) offer a potential alternative by mimicking the pulse-discharge mechanism of biological neurons. These networks trigger pulses only when the membrane potential surpasses a threshold, substantially reducing redundant computations and improving energy efficiency \cite{liu2024spiking}. However, current SNN models face challenges in expressive power and training stability, which limit their effectiveness in handling complex medical diagnostic tasks.

To address the challenges identified, this study introduces a novel hybrid neural network architecture, named FasterSNN, which combines the energy-efficient characteristics of SNN with the expressive power of ANN. FasterSNN incorporates Leaky Integrate-and-Fire (LIF) neurons, the FasterNet \cite{3} model, and a pyramid-structured multi-scale feature fusion attention mechanism. This integration allows for efficient extraction and integration of multi-dimensional feature information from MRI scans, accurately capturing underlying pathological features across different stages of AD, and notably enhancing diagnostic accuracy, particularly during the MCI stage. We validated our model using two authoritative databases, ADNI and AIBL. Experimental results demonstrate that FasterSNN significantly surpasses existing mainstream advanced models in AD classification accuracy. Additionally, FasterSNN shows considerable advantages in energy consumption, requiring only 1.15 Joules. These outcomes highlight FasterSNN's efficiency in resource utilization and its potential for practical deployment. Furthermore, the use of single-modality MRI reduces clinical costs, effectively easing the financial burden on patients and their families, paving the way for the successful integration and application of AI technologies in clinical settings for neurodegenerative diseases. 

\textcolor{black}{
The main contributions of this paper are summarized as follows:
\begin{enumerate}
  \item We propose FasterSNN, a lightweight and interpretable hybrid spiking-artificial neural architecture that leverages LIF-driven sparse computation and region-adaptive 3D convolutions to reduce energy consumption while preserving expressive representational capacity.
  \item We introduce a multi-scale spiking attention mechanism that jointly models channel-wise and spatially distributed pathological patterns in 3D MRI, effectively enabling the network to selectively and adaptively emphasize subtle AD-related structural cues.
  \item We design a hierarchical multi-scale fusion strategy with adaptive and learnable cross-level weighting, allowing the model to effectively integrate substantially relevant coarse-to-fine anatomical information of progressive neurodegeneration.
  \item Extensive experiments on ADNI and AIBL demonstrate that FasterSNN achieves state-of-the-art diagnostic accuracy with only two inference time steps and near 1 J of energy, consistently maintaining strong robustness and high practical deployability.
\end{enumerate}
}

\section{Related Work}
\subsection{Research on AD Prediction Based on Traditional Clinical Medicine}
Traditionally, the early diagnosis of AD has relied on the ADAS-Cog scale and fluid biomarker testing. The ADAS-Cog scale, introduced by Rosen \textit{et al.} in 1984, was designed to address the limitations of existing diagnostic tools at the time. Comprising 12 assessment items, it evaluates various cognitive domains, including memory, language, praxis, and orientation, making it a widely used tool for assessing mild to moderate AD patients. Despite its clinical utility, the scale has significant drawbacks: it struggles to accurately identify very mild or severe dementia and cannot reliably differentiate AD from other forms of dementia, often resulting in misdiagnosis. In addition to the ADAS-Cog scale, biomarker testing plays a key role in AD diagnosis, focusing on pathological hypotheses related to $\beta$-amyloid (A$\beta$) and Tau proteins. Biomarker levels in cerebrospinal fluid (CSF) and blood are measured to aid in diagnostic decisions. Among these, CSF testing is considered the gold standard due to its high sensitivity and specificity, but its utility is constrained by expensive equipment and technical challenges, restricting its use in primary healthcare settings. Blood-based biomarker testing, though more accessible, remains less sensitive compared to CSF analysis. MRI has also become widely adopted in clinical practice due to its lower cost and excellent spatial resolution. MRI is particularly effective at detecting brain atrophy and structural abnormalities associated with AD. However, its accuracy can be compromised by subjective interpretations and overlapping signs of normal aging, such as widened sulci and ventricles. These challenges underscore the need for more objective and automated diagnostic methods, which have increasingly become the focus of contemporary research. 

\subsection{Research on AD Prediction Based on Machine Learning Methods}
In recent years, researchers have increasingly turned to machine learning-based approaches to enhance the accuracy and objectivity of AD diagnoses, addressing the limitations of traditional methods. Magnin \textit{et al.} \cite{20} employed a Support Vector Machine (SVM) to classify MRI data, achieving an impressive accuracy of 94.5\%. However, the study was constrained by a small sample size, and the process required manual delineation of the Region of Interest (ROI), limiting its automation. To address computational complexity and improve accuracy, Bi \textit{et al.} \cite{10} proposed a Random SVM clustering model. While promising, their work was conducted on a small experimental scale, restricting its generalizability. Building on this progress, Lebedev \textit{et al.} \cite{25} applied the Random Forest method on a larger dataset of 410 cases, achieving high sensitivity and specificity (88.5\% and 92.0\%, respectively). Despite these advancements, the model’s construction was highly complex. Ortiz \textit{et al.} \cite{21} introduced a hybrid approach by combining unsupervised segmentation with Learning Vector Quantization (LVQ) to improve classification accuracy. However, this method required extensive preprocessing and was inefficient. Similarly, attempts by Retico \textit{et al.} \cite{22} and Vichianin \textit{et al.} \cite{23} to apply SVM and three-class methods for MCI recognition reported suboptimal accuracy and high energy consumption. Although machine learning methods have shown potential for automating AD diagnosis, challenges such as limited efficiency, insufficient generalization, and the need for more accurate models persist, highlighting the need for further innovation.

\subsection{Research on AD Prediction Based on Existing Neural Networks}
Advances in deep learning technologies, combined with the increasing availability of high-performance computing devices, have significantly improved the efficiency and accuracy of AD diagnosis. Qiu \textit{et al.} \cite{11} pioneered the use of FCN to generate disease probability maps, integrating non-imaging data for classification. Their approach achieved outstanding results across multiple public datasets. Building on this, Kushol \textit{et al.} \cite{12} introduced the ADDFormer model, which leveraged the Vision Transformer (ViT) architecture to enhance classification accuracy by incorporating Fourier features with spatial information. Similarly, Jang \textit{et al.} \cite{14} developed the M3T model, which effectively captured spatial details in 3D MRI data through a combination of CNN and Transformer architectures, yielding superior classification performance. Jiang \textit{et al.} \cite{15} further advanced the field with the AAGN model, integrating transfer learning and anatomical attention mechanisms. This approach improved diagnostic accuracy by offering stronger inductive bias and enhanced interpretability. Additionally, Retico \cite{22} and Ortiz \cite{21} optimized traditional machine learning methods such as SVM and LVQ, achieving higher classification accuracy for MCI recognition.

Despite these advancements, real-world applications of ANN models face notable limitations. Models using 2D slices often suffer from significant spatial information loss, hindering their ability to fully analyze the brain's complex structure in clinical settings. Although models incorporating complete 3D data improve accuracy, they come with considerable drawbacks, including large parameter sizes, prolonged training times, and high computational demands. These challenges are particularly pronounced in resource-limited environments, such as primary healthcare settings, where the deployment of computationally intensive models remains impractical.

\subsection{Research on Classification Based on Existing SNNs}
SNNs, representing the third generation of neural networks, offer substantial and tangible reductions in energy consumption by mimicking the spiking behavior of biologically-inspired and brain-like neurons. Turkson \textit{et al.} \cite{24} were pioneers in applying SNNs to AD classification, achieving improved accuracy through a supervised pretraining method. Similarly, Liu \textit{et al.} \cite{16} developed the Spiking-PhysFormer model, successfully utilizing SNNs in remote photoplethysmography (rPPG), resulting in impressive performance and enhanced energy efficiency. Lin \textit{et al.} \cite{17} introduced the Spike-SLR model, which demonstrated efficient energy consumption in sign language recognition tasks while maintaining high classification accuracy.

Despite these advancements, several critical challenges persist in applying SNNs to AD diagnosis. Firstly, SNNs often struggle with limited expressive power; their feature extraction and generalization capabilities typically fall short compared to traditional CNNs and Transformer models when processing complex medical data like MRI. Secondly, SNNs exhibit poor training stability and convergence, necessitating intricate training strategies that can affect their reliability in clinical settings. Furthermore, research into SNNs for medical imaging, particularly for 3D MRI data, remains relatively sparse, with few mature and stable model architectures currently available. This study aims to address these challenges by combining the energy efficiency of SNNs with the expressive capabilities of traditional neural networks. We propose a hybrid neural network architecture designed to balance accuracy and energy efficiency, thereby overcoming the limitations of existing methods in clinical applications.

\section{Methodology}

\begin{figure}[H]
\centering
\includegraphics[width=\textwidth]{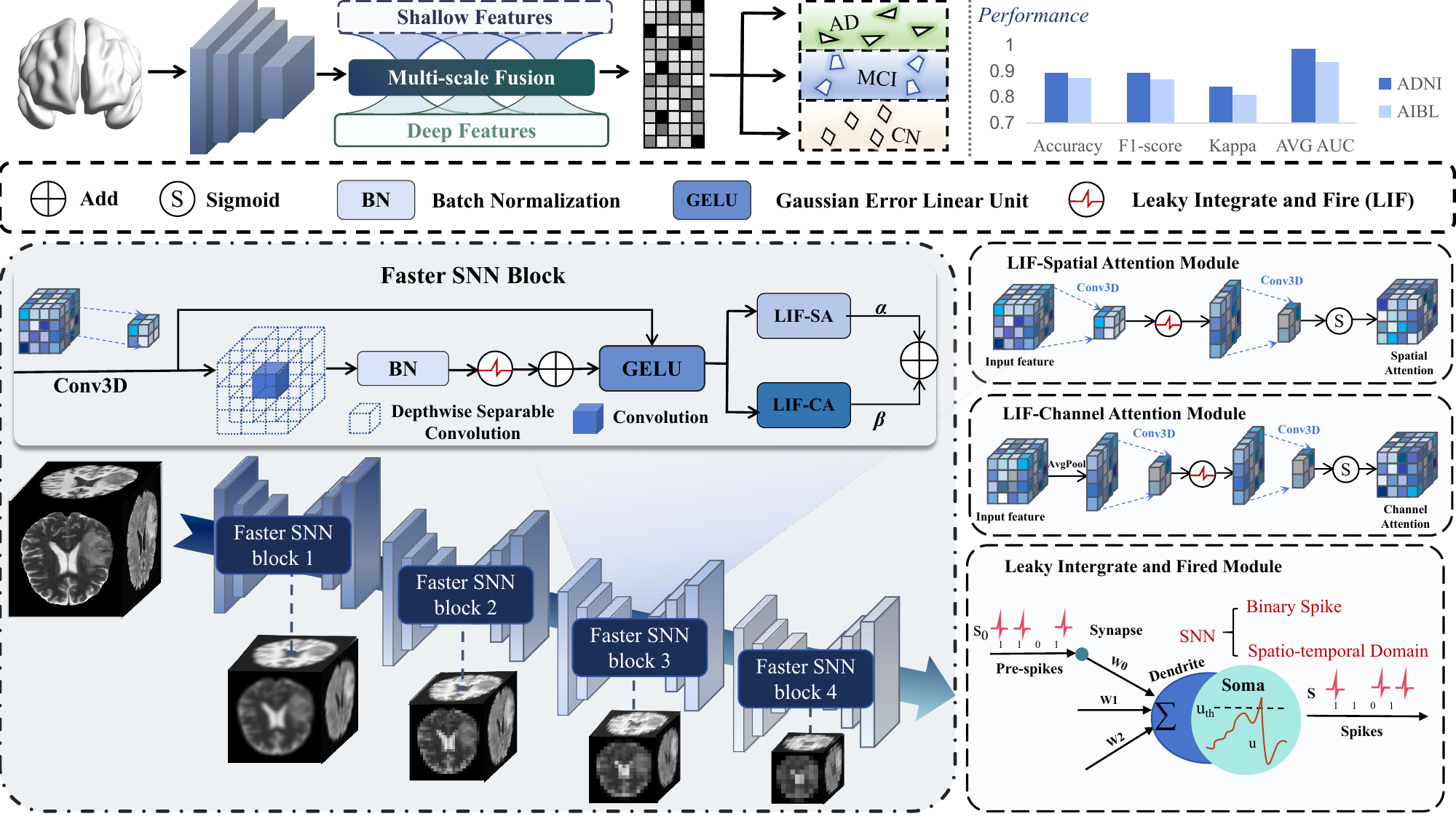}
\caption{\textbf{Overall architecture of the FasterSNN.} It stacks multiple Faster SNN blocks: the first three blocks include Depthwise Separable Convolution, Batch Normalization, Gaussian Error Linear Unit, and SWA modules, while the last block excludes the SWA module. The SWA module integrates two types of attention mechanisms: the LIF Spatial Attention and the LIF Channel Attention.}
\label{model}
\end{figure}
 This study proposes a model to classify 3D Alzheimer's MRI data using a SNN based on multi-scale feature fusion, as illustrated in Fig.~\ref{model}. The model consists of four primary components: LIF neurons, the Spike-weighted Attention (SWA) module, the Faster SNN Blocks, and the Multi-Scale Feature Fusion (MSF) module. Initially, the model employs LIF neurons with a membrane potential update mechanism to carry out sparse computations that are biologically interpretable. The 3D MRI data first passes through a convolutional layer, which sets the stage for further processing through four feature extraction blocks. These blocks utilize Faster SNN Blocks as their core network layers. Within this module, a region-adaptive convolution strategy is implemented, applying different convolution methods to distinct regions of the feature map. The features extracted are then concatenated to enrich the data representation. The SWA module enhances the model by introducing trainable parameters within each Faster SNN Block, enabling dynamic weighting through pulse channel and spatial attention mechanisms. The overall network structure is shaped by the MSF module, which constructs a four-level pyramid architecture. This setup includes cross-layer connections that facilitate feature reuse. Learnable parameters are employed to weight and aggregate the output of each layer, enhancing feature integration. Finally, the model performs temporal averaging on the fused features before feeding them into the classifier, ensuring accurate classification outcomes.

\subsection{Leaky Integrate-and-Fire Neuron}
In classification tasks, traditional ANN models face significant computational and energy demands due to the high-resolution, three-dimensional voxel structure of 3D MRI data. SNN models offer a more biologically plausible and energy-efficient alternative by mimicking the membrane potential updates and spike emission mechanisms observed in biological neurons. This event-driven approach enables sparse information transmission, resulting in reduced computational costs. This study adopts the LIF neuron model, as described by Wu \textit{et al.} \cite{26}. In this model, a neuron's membrane potential is updated at each time step by adding the current input while simultaneously applying a decay factor to the membrane potential from the previous step. When the membrane potential exceeds a predefined threshold, the neuron generates a spike and resets its potential. This process allows the model to more closely resemble the operation of biological neural systems while improving its efficiency in representing complex 3D MRI data. The sparsity of LIF neurons further enhances the model's ability to identify and focus on critical patterns within the data, reducing the likelihood of overfitting. This characteristic strengthens the reliability of subsequent modules for feature extraction and classification. The membrane potential update and spike emission mechanism of LIF neurons can be formally and rigorously expressed mathematically as follows:
\begin{equation}
\begin{gathered}
    V_t = \lambda V_{t-1} + \frac{1}{T} \sum_{i=1}^{T} \mathbf{x}_t^{(i)},S_t = 
    \begin{cases}
        1 & \text{if } V_t \geq V_{th} \\
        0 & \text{otherwise},
    \end{cases}
\end{gathered}
\end{equation}
where \( V_t \) denotes the membrane potential at each time step \( t \). The parameter \( T \) defines the time step, which by default is set to a value of 2. The decay rate of the membrane potential from the previous time step is regulated by \( \lambda \), with an initial setting of 0.9 to ensure gradual attenuation. The threshold for spike emission is represented by \( V_{th} \), initially set to 1. Furthermore, \( S_t \) marks the binary state of spike emission.

\subsection{Spiking Weighted Attention Module}
Pathological features in 3D Alzheimer's MRI data are often faint and sparse, making it crucial to improve classification accuracy by effectively extracting key information and minimizing irrelevant voxel interference. This study employs LIF neurons to develop the SWA module. The SWA module integrates the dynamic modeling capabilities of attention mechanisms with the temporal sparsity characteristic of spiking neurons, enhancing the network's ability to identify important lesion areas. This is achieved through fine-grained modeling and weighted modifications of the input feature map using parallel pulse channel attention and pulse spatial attention structures.

The SWA module begins by aggregating spatial information in the pulse channel attention branch. It accomplishes this by applying global average pooling (GAP) to the input feature map, resulting in a feature representation with dimensions \( (B, C, 1, 1, 1) \). To reduce computational costs, the number of channels is condensed to one-fourth of their original size using a \( 1 \times 1 \times 1 \) convolution. Subsequently, the LIF neuron is utilized for spike discretization, enhancing information sparsity. Another \( 1 \times 1 \times 1 \) convolution then restores the channels to their initial size, and the channel weight map is produced through the Sigmoid activation function. Similarly, the pulse spatial attention branch performs analogous operations on the input feature map, thereby capturing rich contextual dependencies and ultimately generating a more informative attention distribution across spatial dimensions.

To facilitate adaptive adjustment of the dual-path fusion, the SWA module includes two learnable parameters that regulate the proportion of the two types of attention in the final fusion, dynamically balancing their contribution. The specific formula governing these processes is as follows:
\begin{equation}
\begin{gathered}
    \bm{W}_c = \sigma\left(\text{LIF}\left(\text{Conv}\left(\text{GAP}(\bm{X})\right)\right)\right),\\[6pt]
    \bm{W}_s = \sigma\left(\text{LIF}\left(\text{Conv}\left(\text{Conv}(\bm{X})\right)\right)\right),\\[6pt]
    \bm{X}_{\text{out}} = \bm{X} \odot (\alpha \bm{W}_c + \beta \bm{W}_s),
\end{gathered}
\end{equation}
where \(\sigma\) denotes the widely used Sigmoid activation function, the parameters \(\alpha\) and \(\beta\) represent the learnable weights assigned to channel attention and spatial attention, respectively, each initially set at 0.5, and the symbol \(\odot\) indicates the standard element-wise multiplication operation.

\subsection{Faster SNN Block}
To enhance computational efficiency in processing 3D high-resolution MRI data, this study builds upon the region-adaptive concept of FasterNet and integrates it with LIF neurons to introduce the Faster SNN Block. This module partitions the input feature map into core and edge regions and applies distinct convolution techniques to each. For the core region, standard \( 3 \times 3 \times 3 \) convolution is used to extract comprehensive feature information, while the edge region employs depthwise separable convolution to reduce computational load. This approach balances feature extraction capacity with computational efficiency, preserving the receptive field while optimizing processing speed.

To maintain spatial consistency in the final concatenated feature map, the Faster SNN Block incorporates a spatial masking method. Using a mask matrix \(\mathbf{M}\), it identifies and separates the core and edge regions. Additionally, residual connections are introduced in the main path to improve training stability and prevent vanishing gradient issues. Sparse features generated by the LIF neurons are added to the residual path to strengthen feature learning and representation. These design elements allow the Faster SNN Block to produce high-quality feature representations, supporting subsequent multi-scale fusion and classification tasks while significantly improving computational efficiency and accuracy. The specific mathematical framework for the Faster SNN Block is expressed as follows:
\begin{equation}
\begin{gathered}
    \bm{F}_{\text{center}} = \text{Conv}_{3\times3\times3}\left(\bm{X}[:, :, h_s:h_e, w_s:w_e, d_s:d_e]\right),\\[6pt]
    \bm{F}_{\text{edge}} = \text{Conv}_{1\times1\times1}\left(\text{Conv}_{\text{depthwise}}(\bm{X})\right),\\[6pt]
    M_{i,j,k} =
    \begin{cases}
        1 & \text{if } (i,j,k) \in \text{center region} \\
        0 & \text{otherwise},
    \end{cases}\\
    \bm{F}_{\text{out}} = \bm{M} \odot \bm{F}_{\text{center}} + \bm{M} \odot \bm{F}_{\text{edge}},  
    \bm{Y} = \bm{X} + \text{LIF}\left(\text{BN}(\bm{F}_{\text{out}})\right),
\end{gathered}
\end{equation}
where $h_s:h_e$, $w_s:w_e$, $d_s:d_e$ represent the starting and ending indices for the height, width, and depth, respectively, and $\text{Conv}_{\text{depthwise}}$ denotes depthwise separable convolution.

\subsection{Multi-Scale Feature Fusion Module}
AD is a chronic and progressive neurodegenerative disorder. Its MRI features show distinct multi-scale distribution characteristics. Single-scale characteristics alone cannot fully portray the imaging patterns of AD. This is because AD involves both macroscopic brain region atrophy and other lesions, as well as microscopic neuronal degeneration.

To address this, we designed the MSF module in this study. The MSF aims to comprehensively analyze multi-scale feature information across different layers. The module is structured as a four-level pyramid. The number of channels gradually expands from 64 to 128, then to 256, and finally reaches 512. After each layer produces an output, a \(1 \times 1\) convolution is used to adjust the output to 64 channels. The outputs of the first three layers are downsampled to the same size through Max Pooling. Then, they are added to the fourth layer's feature map, enabling spatially aligned fusion. MSF further enhances cross-layer fusion by using weighted summation to integrate multi-scale features. Specifically, after convolution and pooling operations, each layer's feature map is multiplied by a learnable weight. This allows for dynamic adjustment of the contributions from different layers. These weights are automatically optimized during training to make the most of the discriminative power of each layer.

Finally, the model conducts temporal averaging on the fused feature maps generated at each individual time step, thereby integrating complementary temporal dynamics into a unified representation. This operation yields the final, temporally enriched fused feature map, which is subsequently forwarded to the classifier for precise and reliable Alzheimer's image classification. The specific mathematical formulations are defined as follows:
\begin{equation}
\begin{gathered}
    \bm{F}_{\text{fusion}} = \sum_{i=1}^{N} w_i \cdot \text{Pool}\left(\text{Conv}_{1\times1}(\bm{X}_{\text{out}})\right),\\[6pt]
    \bm{F}_{\text{final}} = \frac{1}{T} \sum_{t=1}^{T} \bm{F}_{\text{fusion}}^{(t)},
\end{gathered}
\end{equation}
where $w_i$ represents the weight of the feature map at the $i$-th layer, with an initial value of 1.0.

\section{Experiments}
\subsection{Datasets}
\subsubsection{ADNI Dataset}
In this research, we combined two data sources from the ADNI \cite{weiner2015impact} to create a neuroimaging dataset. This dataset consists of three categories: AD, MCI, and Cognitively Normal (CN). The first data source was obtained from the Kaggle public platform. It contains 424 images that have been processed into the NIfTI format. These images have a resolution of 256×256×166. Among them, 168 images are from AD patients and 254 are from CN controls. The second data source was directly retrieved from the ADNI platform by our research team. The DICOM data was automatically converted and sequenced to form NIfTI images. These images have a spatial resolution of 256×256×92, and we selected 283 MCI cases from this source. The entire ADNI dataset was stratified and sampled at an 8:2 ratio. This was done to create training and testing sets, which are crucial for evaluating the performance of the model.

\subsubsection{AIBL Dataset}
This study acquired and processed 682 three-dimensional neuroimaging cases from the Australian Imaging, Biomarker, and Lifestyle (AIBL \cite{fowler2021fifteen} platform, including 219 AD cases, 263 MCI cases, and 200 CN controls, with an image spatial resolution of 160×240×256. For the AIBL dataset, a stratified and random sampling approach was used. The data was sampled at a 7:3 ratio to form the training and testing sets. This sampling method helps in getting a representative subset of the data for model training and evaluation.

\subsection{Data Preprocessing}
To make the data suitable for the three - dimensional image classification task and to enhance the efficiency of model training, we took several steps in data preprocessing. First, all images were resampled to a uniform size of 64×64×64. Then, Z - score normalization was performed on the images. In addition, we generated a temporal sequence with Gaussian noise based on the images. The output of this process was a PyTorch tensor with a shape of (Time\_Steps, 64, 64, 64). These preprocessing steps help to standardize the data and add an element of variability that can potentially improve the model's generalization ability. \textcolor{black}{These preprocessing steps help standardize the data and minimize the impact of source-related discrepancies on the model.}

\subsection{Evaluation indicators}
In this study, we used a multidimensional evaluation system. This system allows us to analyze the model from two important aspects: classification performance and computational efficiency. By considering both aspects, we can get a more complete picture of the model's overall performance.

\subsubsection{Classification performance}
When it comes to measuring classification performance, we used standard metrics. These include accuracy, precision, recall, and F1 - score. We also considered the Kappa coefficient and the average area under the curve (AUC) value. These indicators are especially important because our datasets may be imbalanced.

\subsubsection{Energy consumption}
This study presents a comprehensive evaluation framework designed to assess computational efficiency through three key dimensions: energy consumption, complexity of parameter scale, and training time cost. By integrating these factors, the approach provides a thorough analysis of a model's resource utilization efficiency. 
\textcolor{black}{%
Regarding the calculation of model energy consumption, we note that obtaining accurate energy measurements for SNNs on conventional GPUs is infeasible, as their intrinsic energy characteristics can be quantified only with specialized neuromorphic hardware. Consequently, we adopted the methodology established by Song et al. \cite{song2024learning} to theoretically estimate the model's energy consumption during the inference of a single image. The specific calculation formula is presented as follows:
}
\begin{equation}
\begin{gathered}
\begin{aligned}
e_{\text{spike}} &= 77 \times 10^{-15} \, \text{J}, & e_{\text{mac}} &= 12.5 \times 10^{-12} \, \text{J}, & \text{Sparsity} &= 1 - \frac{N_{\text{active}}}{N_{\text{total}}},
\end{aligned}\\[6pt]
\text{MACs} = C_{\text{in}} \times C_{\text{out}} \times k^3 \times W_{\text{out}} \times H_{\text{out}} \times D_{\text{out}},\\[6pt]
\text{Energy} = (\text{MACs} \times e_{\text{mac}} \times (1 - \text{Sparsity})) + \text{Spikes} \times e_{\text{spike}},
\end{gathered}
\end{equation}
\textcolor{black}{
where $N_{\text{active}}$ and $N_{\text{total}}$ denote the number of active neurons and total neurons, respectively, and Sparsity represents the neuronal sparsity.}

\subsubsection{Parameter amount calculation}
The process of calculating the total number of model parameters involves summing up all the trainable parameters, which include both weights and biases from each layer within the network. This total is then expressed in millions (M). Specifically, the formulation can be represented as follows:
\begin{equation}
Params = \sum_{l=1}^{L} \left( W_l^{\text{shape}} + b_l^{\text{shape}} \right),
\end{equation}
where \( L \) denotes the total number of layers in the network. The term \( W_l^{\text{shape}} \) corresponds to the product of the dimensions of the weight parameters in the \( l \)-th layer, while \( b_l^{\text{shape}} \) represents the product of the dimensions of the bias parameters in the same layer.

\subsubsection{Time Calculation}
\textcolor{black}{%
To precisely measure the training duration for each model, we recorded a timestamp at the start of training and another at its end. The time difference between these two timestamps reflects the duration of a single epoch. The specific formulation for this calculation is provided below:
}

\begin{equation}
Time_{\text{train}} = Time_{\text{end}} - Time_{\text{start}},
\end{equation}
\textcolor{black}{%
where $Time_{start}$ denotes the timestamp at the beginning of the current epoch, and $Time_{end}$ represents the timestamp at the end of the current epoch.
}

\subsection{Training Methodology}
\textcolor{black}{%
During training, we used the standard cross-entropy loss function. To obtain the overall loss, the output probabilities were averaged across all timesteps , thereby enhancing learning stability along the temporal dimension. The network parameters were then optimized end-to-end using the Spatio-Temporal Backpropagation (STBP) algorithm. This algorithm unfolds the temporal dynamics of LIF neurons. It propagates error signals backward incrementally, thereby effectively preserving critical temporal coding and dynamic characteristics. Given the inherent non-differentiability of the neuronal spiking function, we employed a rectangular approximation in the form of a Surrogate Gradient (SG) function for gradient computation. This specific form has demonstrated superior gradient smoothness and training stability in practical implementation. To further ensure robust training, we applied global gradient clipping to mitigate gradient explosion and incorporated a cosine annealing learning rate scheduler to improve overall convergence. The specific calculation formulas are as follows:
}

\begin{equation}
\begin{gathered}
\begin{aligned}
\mathcal{L}_{\text{CE}} &= \mathrm{CE}(Y_t, Y_{tue}),
s =
\begin{cases}
a, & \text{if } |V_{\text{IV}} - V_h| < 0.5,\\
0, & \text{otherwise}.
\end{cases}
\end{aligned}
\end{gathered}
\end{equation}

\textcolor{black}{%
\\where $CE(\cdot)$ denotes the cross-entropy computation, $y_t$ represents the predicted probability distribution at timestep $t$, $y_{true}$ denotes the ground-truth label, and $V_{th}$ is the neuronal firing threshold, which was set to $1.0$ by default.
}

\subsection{Experimental Details}
In this experiment, we utilized the PyTorch 2.4.1 deep learning framework on a hardware platform equipped with an NVIDIA Tesla V100 GPU and CUDA 11.8. To enhance computational efficiency, we implemented a mixed-precision training strategy. For data processing, both the training and test sets were configured with a batch size of 16. During training, a dynamic random shuffle strategy was employed to improve model generalizability, complemented by a 4-thread parallel data loader to accelerate preprocessing. The Adam optimizer was chosen for model optimization, with the initial learning rate set to \( 1 \times 10^{-3} \). To further regularize the model, a weight decay coefficient of \( 1 \times 10^{-3} \) was applied. Additionally, a ReduceLROnPlateau dynamic learning rate adjustment strategy was implemented: if validation accuracy failed to improve after three consecutive epochs, the learning rate was halved to mitigate the risk of converging to local optima. Global gradient clipping was also utilized to ensure stable training dynamics. For models incorporating LIF neurons, key parameters were configured to emulate the dynamic behavior of biological neurons. Specifically, the initial time step was set to 2, the spike firing threshold to 1, and the decay coefficient to 0.9. These adjustments were made to align the model's behavior with the temporal characteristics observed in neural systems.
\textcolor{black}{%
To ensure a fair and rigorous comparison, all baseline models involved in this study were retrained and evaluated using the above identical preprocessing pipeline, data loader, optimizer settings, and hardware platform.
}

\subsection{Comparative Experiments}
\subsubsection{Quantitative Evaluation of Performance}

\textcolor{black}{%
To enhance the stability of the comparative model results and minimize the uncertainty introduced by the random seed, this study employed a five-fold cross-validation strategy. During this process, the overall training data was partitioned into training and validation sets at a 4:1 ratio. The weights that yielded the best performance on the validation set were loaded for each model across the cross-validation folds. Subsequently, the final model performance was evaluated on a dedicated, independent test set.
}\\

\begin{figure}[t]
\centering
\includegraphics[width=\linewidth]{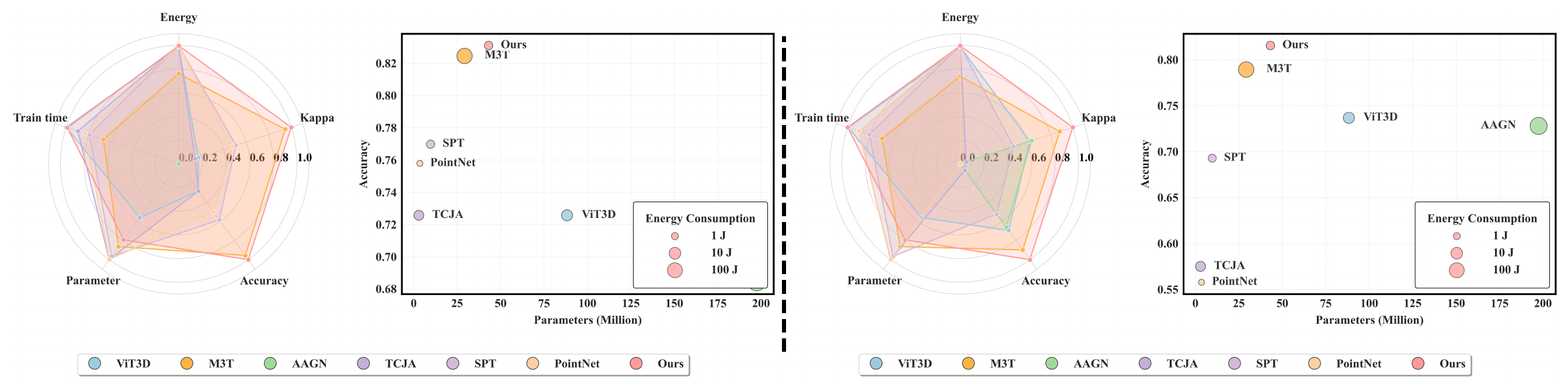}
\caption{Performance analysis of various models on the ADNI (left) and AIBL (right) datasets. The radar charts illustrate models' performance on five metrics: energy, training time, number of parameters, accuracy, and Kappa. Each axis represents one metric, normalized between 0 and 1, with spoke length indicating the relative score. The bubble charts compare models based on energy, parameters, and accuracy. The x-axis shows the number of parameters in millions, the y-axis indicates accuracy, and bubble size reflects energy consumption in joules.
}                 
\label{rose}
\end{figure}

\begin{figure}[t]
\centering
\includegraphics[width=\textwidth]{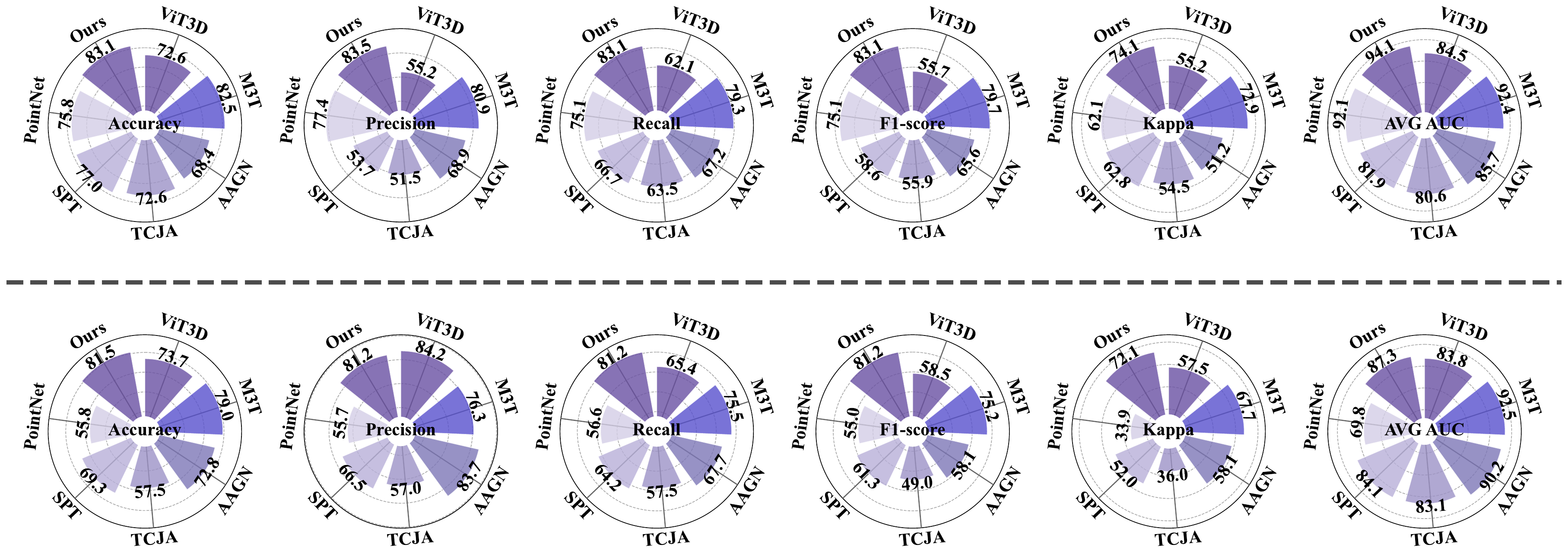}  
\caption{Classification performance metrics of various models on the ADNI and AIBL datasets. The upper half of the rose plot presents six categorical metrics: accuracy, precision, recall, F1-score, Kappa, and average AUC on the ADNI dataset. The lower half shows the same metrics on the AIBL dataset. All models are clearly labeled and color-coded to facilitate visual comparison.}                   
\label{rodar}
\end{figure}

\begin{table}[H]
\centering
\renewcommand{\tabularxcolumn}[1]{>{\centering\arraybackslash}m{#1}}
\tiny
\renewcommand{\arraystretch}{1.2}
\caption{\color{black}Performance metrics obtained through five-fold cross-validation on the ADNI dataset. Red highlights the best results showing highest metrics and lowest resource usage, with blue highlighting the second-best.}
\label{tab:ADNI_five_folds}
\color{black}
\setlength{\arrayrulewidth}{0.5pt}
\begin{tabularx}{\textwidth}{@{}l*{10}{X}@{}}
\hline
& \multicolumn{6}{c}{Metrics ($\uparrow$)} & \multicolumn{2}{c}{Resources ($\downarrow$)} & Step ($\downarrow$) \\
\cline{2-10}
Model & Accuracy & Precision & Recall & F1-score & Kappa & AUC & Train Time (s) & Parameter (M) & Time Step \\
\hline
ResNet50 \cite{30}  & 0.8060 $\pm$ 0.0286 & 0.8073 $\pm$ 0.0335 & 0.7636 $\pm$ 0.0489 & 0.7514 $\pm$ 0.0671 & 0.6975 $\pm$ 0.0506 & 0.9064 $\pm$ 0.0293 & 23 & 46.16  & -- \\
ResNet101 \cite{30} & 0.8172 $\pm$ 0.0527 & 0.8169 $\pm$ 0.0534 & \textcolor{red}{0.8121} $\pm$ 0.0605 & \textcolor{red}{0.8121} $\pm$ 0.0584 & 0.7202 $\pm$ 0.0847 & \textcolor{blue}{0.9152} $\pm$ 0.0313 & 28 & 85.21  & -- \\
FCN \cite{11}       & 0.8149 $\pm$ 0.0459 & 0.7970 $\pm$ 0.0450 & 0.7710 $\pm$ 0.0616 & 0.7678 $\pm$ 0.0649 & 0.7108 $\pm$ 0.0739 & 0.9051 $\pm$ 0.0388 & 28 & 10.09  & -- \\
ViT \cite{33}       & 0.7478 $\pm$ 0.0271 & 0.6270 $\pm$ 0.1390 & 0.6617 $\pm$ 0.0256 & 0.5928 $\pm$ 0.0392 & 0.5963 $\pm$ 0.0402 & 0.8496 $\pm$ 0.0226 & \textcolor{blue}{17} & 88.26  & -- \\
M3T \cite{14}       & \textcolor{blue}{0.8236} $\pm$ 0.0298 & \textcolor{blue}{0.8216} $\pm$ 0.0214 & 0.7732 $\pm$ 0.0612 & 0.7603 $\pm$ 0.0825 & \textcolor{blue}{0.7227} $\pm$ 0.0541 & 0.9063 $\pm$ 0.0390 & 106 & 29.23 & -- \\
AAGN \cite{15}      & 0.7080 $\pm$ 0.0280 & 0.7126 $\pm$ 0.0337 & 0.7081 $\pm$ 0.0337 & 0.7008 $\pm$ 0.0357 & 0.5585 $\pm$ 0.0465 & 0.8602 $\pm$ 0.0105 & 287 & 197.51 & -- \\
VGGSNN \cite{31}    & 0.7054 $\pm$ 0.0424 & 0.6043 $\pm$ 0.0983 & 0.7054 $\pm$ 0.0424 & 0.6435 $\pm$ 0.0713 & 0.5391 $\pm$ 0.0583 & 0.8786 $\pm$ 0.0196 & \textcolor{red}{15} & 10.08 & 16 \\
ResSNN \cite{36}    & 0.7249 $\pm$ 0.0339 & 0.6472 $\pm$ 0.1112 & 0.7249 $\pm$ 0.0339 & 0.6620 $\pm$ 0.0664 & 0.5650 $\pm$ 0.0588 & 0.8643 $\pm$ 0.0248 & \textcolor{blue}{17} & 5.85 & 4 \\
TCJA \cite{32}      & 0.7302 $\pm$ 0.0226 & 0.5073 $\pm$ 0.0175 & 0.6418 $\pm$ 0.0123 & 0.5586 $\pm$ 0.0139 & 0.5671 $\pm$ 0.0300 & 0.8373 $\pm$ 0.0311 & 45 & \textcolor{red}{2.87} & 16 \\
SPT \cite{35}       & 0.7567 $\pm$ 0.0124 & 0.6325 $\pm$ 0.1281 & 0.6742 $\pm$ 0.0212 & 0.6098 $\pm$ 0.0483 & 0.6126 $\pm$ 0.0155 & 0.8433 $\pm$ 0.0188 & 73 & 9.64 & 2 \\
PointNet \cite{34}  & 0.5907 $\pm$ 0.1692 & 0.6188 $\pm$ 0.1502 & 0.5911 $\pm$ 0.1721 & 0.5676 $\pm$ 0.1914 & 0.3797 $\pm$ 0.2547 & 0.7409 $\pm$ 0.1478 & 62 & \textcolor{blue}{3.47} & 16 \\
Ours                & \textcolor{red}{0.8325} $\pm$ 0.0276 & \textcolor{red}{0.8247} $\pm$ 0.0232 & \textcolor{blue}{0.7886} $\pm$ 0.0355 & \textcolor{blue}{0.7912} $\pm$ 0.0357 & \textcolor{red}{0.7381} $\pm$ 0.0440 & \textcolor{red}{0.9294} $\pm$ 0.0153 & 20 & 43.11 & 2 \\
\hline
\end{tabularx}
\end{table}

\begin{table}[H]
\centering
\renewcommand{\tabularxcolumn}[1]{>{\centering\arraybackslash}m{#1}}
\tiny
\renewcommand{\arraystretch}{1.2}
\caption{\color{black}Performance metrics obtained through five-fold cross-validation on the AIBL dataset. Red highlights the best results showing highest metrics and lowest resource usage, with blue highlighting the second-best.}
\label{tab:AIBL_five_folds}
\color{black}
\setlength{\arrayrulewidth}{0.5pt}
\begin{tabularx}{\textwidth}{@{}l*{10}{X}@{}}
\hline
& \multicolumn{6}{c}{Metrics ($\uparrow$)} & \multicolumn{2}{c}{Resources ($\downarrow$)} & Step ($\downarrow$) \\
\cline{2-10}
Model & Accuracy & Precision & Recall & F1-score & Kappa & AUC & Train Time (s) & Parameter (M) & Time Step \\
\hline
ResNet50 \cite{30}  & 0.7655 $\pm$ 0.0267 & 0.7746 $\pm$ 0.1260 & 0.6870 $\pm$ 0.0343 & 0.6329 $\pm$ 0.0619 & 0.6263 $\pm$ 0.0423 & 0.8684 $\pm$ 0.0330 & 21 & 46.16 & -- \\
ResNet101 \cite{30} & 0.7832 $\pm$ 0.0326 & \textcolor{blue}{0.7930} $\pm$ 0.0379 & 0.7233 $\pm$ 0.0272 & 0.7014 $\pm$ 0.0613 & 0.6599 $\pm$ 0.0481 & 0.8823 $\pm$ 0.0130 & 28 & 85.21 & -- \\
FCN \cite{11}       & 0.7779 $\pm$ 0.0350 & 0.7490 $\pm$ 0.1413 & 0.7048 $\pm$ 0.0269 & 0.6656 $\pm$ 0.0625 & 0.6483 $\pm$ 0.0502 & 0.8731 $\pm$ 0.0254 & 27 & 10.09 & -- \\
ViT \cite{33}       & 0.7284 $\pm$ 0.0212 & 0.6260 $\pm$ 0.1316 & 0.6470 $\pm$ 0.0158 & 0.5859 $\pm$ 0.0392 & 0.5672 $\pm$ 0.0300 & 0.8451 $\pm$ 0.0235 & 19 & 88.26 & -- \\
M3T \cite{14}       & 0.7848 $\pm$ 0.0120 & 0.7830 $\pm$ 0.0504 & 0.7231 $\pm$ 0.0275 & 0.7066 $\pm$ 0.0470 & 0.6608 $\pm$ 0.0222 & 0.8930 $\pm$ 0.0271 & 93 & 29.23 & -- \\
AAGN \cite{15}      & \textcolor{blue}{0.7885} $\pm$ 0.0430 & \textcolor{red}{0.7973} $\pm$ 0.0316 & \textcolor{blue}{0.7350} $\pm$ 0.0566 & \textcolor{blue}{0.7099} $\pm$ 0.0858 & \textcolor{blue}{0.6685} $\pm$ 0.0680 & \textcolor{blue}{0.9002} $\pm$ 0.0308 & 265 & 197.51 & -- \\
VGGSNN \cite{31}    & 0.6948 $\pm$ 0.0307 & 0.6124 $\pm$ 0.1135 & 0.6948 $\pm$ 0.0307 & 0.6119 $\pm$ 0.0291 & 0.5166 $\pm$ 0.0450 & 0.8674 $\pm$ 0.0140 & \textcolor{red}{14} & 10.08 & 16 \\
ResSNN \cite{36}    & 0.7301 $\pm$ 0.0383 & 0.6786 $\pm$ 0.1274 & 0.7301 $\pm$ 0.0383 & 0.6882 $\pm$ 0.0778 & 0.5778 $\pm$ 0.0672 & 0.8860 $\pm$ 0.0209 & \textcolor{blue}{15} & 5.85 & 4 \\
TCJA \cite{32}      & 0.6156 $\pm$ 0.0520 & 0.5403 $\pm$ 0.0856 & 0.6156 $\pm$ 0.0520 & 0.5426 $\pm$ 0.0701 & 0.4154 $\pm$ 0.0549 & 0.8404 $\pm$ 0.0145 & 42 & \textcolor{red}{2.87} & 16 \\
SPT \cite{35}       & 0.7196 $\pm$ 0.0350 & 0.5606 $\pm$ 0.0853 & 0.6349 $\pm$ 0.0343 & 0.5748 $\pm$ 0.0434 & 0.5495 $\pm$ 0.0577 & 0.8336 $\pm$ 0.0297 & 65 & 9.64 & 2 \\
PointNet \cite{34}  & 0.5983 $\pm$ 0.1308 & 0.6114 $\pm$ 0.1215 & 0.5880 $\pm$ 0.1362 & 0.5775 $\pm$ 0.1484 & 0.3844 $\pm$ 0.2050 & 0.7343 $\pm$ 0.1153 & 42 & \textcolor{blue}{3.47} & 16 \\
Ours                & \textcolor{red}{0.8099} $\pm$ 0.0494 & 0.7500 $\pm$ 0.1175 & \textcolor{red}{0.7522} $\pm$ 0.0849 & \textcolor{red}{0.7272} $\pm$ 0.1123 & \textcolor{red}{0.6980} $\pm$ 0.0921 & \textcolor{red}{0.9088} $\pm$ 0.0384 & 17 & 43.11 & 2 \\
\hline
\end{tabularx}
\end{table}

\normalcolor 

\begin{table}[t]
\centering
\renewcommand{\tabularxcolumn}[1]{>{\centering\arraybackslash}m{#1}}
\tiny
\renewcommand{\arraystretch}{1.2}
\caption{\color{black}Test results using the best weights obtained through five-fold cross-validation for each model on the ADNI dataset.
Values in red represent optimal performance across all metrics and resources, while values blue represent the second-best outcomes.}
\label{tab:ADNI_test}
\color{black}
\setlength{\arrayrulewidth}{0.5pt}
\begin{tabularx}{\textwidth}{@{}l*{10}{X}@{}}
\hline
& \multicolumn{6}{c}{Metrics ($\uparrow$)} & \multicolumn{2}{c}{Resources ($\downarrow$)} & Step ($\downarrow$) \\
\cline{2-10}
Model & Accuracy & Precision & Recall & F1-score & Kappa & AUC & Energy (J) & Parameter (M) & Time Step \\
\hline
ResNet50 \cite{30}   & 0.8230 & 0.7938 & 0.7650 & 0.7655 & 0.7223 & 0.9207 & 6.38  & 46.16 & -- \\
ResNet101 \cite{30}  & 0.8233 & \textcolor{blue}{0.8107} & \textcolor{blue}{0.8070} & \textcolor{blue}{0.8075} & \textcolor{blue}{0.7345} & \textcolor{blue}{0.9378} & 9.54  & 85.21 & -- \\
FCN \cite{11}        & 0.8230 & 0.8010 & 0.8093 & 0.7990 & 0.7308 & 0.9264 & 14.77 & 10.09 & -- \\
ViT \cite{33}        & 0.7257 & 0.5524 & 0.6212 & 0.5572 & 0.5518 & 0.8445 & 7.14  & 88.26 & -- \\
M3T \cite{14}        & \textcolor{blue}{0.8246} & 0.8090 & 0.7934 & 0.7972 & 0.7287 & 0.9236 & 148.64 & 29.23 & -- \\
AAGN \cite{15}       & 0.6842 & 0.6892 & 0.6718 & 0.6557 & 0.5123 & 0.8568 & 622.29 & 197.51 & -- \\
ResSNN \cite{36}     & 0.6903 & 0.5228 & 0.6903 & 0.5945 & 0.5037 & 0.8476 & \textcolor{blue}{1.64} & 5.85 & 4 \\
VGGSNN \cite{31}     & 0.6930 & 0.5273 & 0.6930 & 0.5984 & 0.5073 & 0.8667 & 1.76  & 10.08 & 16 \\
TCJA \cite{32}       & 0.7257 & 0.5149 & 0.6349 & 0.5593 & 0.5455 & 0.8060 & 3.61  & \textcolor{red}{2.87} & 16 \\
SPT \cite{35}        & 0.7699 & 0.5373 & 0.6667 & 0.5864 & 0.6284 & 0.8194 & 1.67  & 9.64 & 2 \\
PointNet \cite{34}   & 0.7579 & 0.7737 & 0.7508 & 0.7512 & 0.6214 & 0.9209 & \textcolor{red}{0.67} & \textcolor{blue}{3.47} & 16 \\
Ours                 & \textcolor{red}{0.8310} & \textcolor{red}{0.8346} & \textcolor{red}{0.8310} & \textcolor{red}{0.8312} & \textcolor{red}{0.7408} & \textcolor{red}{0.9406} & 1.74  & 43.11 & 2 \\
\hline
\end{tabularx}
\end{table}

\begin{table}[t]
\centering
\renewcommand{\tabularxcolumn}[1]{>{\centering\arraybackslash}m{#1}}
\tiny
\renewcommand{\arraystretch}{1.2}
\caption{\color{black}Test results using the best weights obtained through five-fold cross-validation for each model on the AIBL dataset.
Values in red represent optimal performance across all metrics and resources, while values blue represent the second-best outcomes.}
\label{tab:AIBL_test}
\color{black}
\setlength{\arrayrulewidth}{0.5pt}
\begin{tabularx}{\textwidth}{@{}l*{10}{X}@{}}
\hline
& \multicolumn{6}{c}{Metrics ($\uparrow$)} & \multicolumn{2}{c}{Resources ($\downarrow$)} & Step ($\downarrow$) \\
\cline{2-10}
Model & Accuracy & Precision & Recall & F1-score & Kappa & AUC & Energy (J) & Parameter (M) & Time Step \\
\hline
ResNet50 \cite{30}  & \textcolor{blue}{0.8053} & 0.7652 & \textcolor{blue}{0.7537} & \textcolor{blue}{0.7547} & 0.6968 & 0.9088 & 6.92  & 46.16 & -- \\
ResNet101 \cite{30} & 0.7895 & 0.8014 & 0.7301 & 0.7241 & 0.6665 & 0.8777 & 10.12 & 85.21 & -- \\
FCN \cite{11}       & 0.7965 & \textcolor{red}{0.8802} & 0.7051 & 0.6626 & \textcolor{blue}{0.6731} & 0.8870 & 13.85 & 10.09 & -- \\
ViT \cite{33}       & 0.7368 & \textcolor{blue}{0.8422} & 0.6542 & 0.5854 & 0.5754 & 0.8385 & 7.50  & 88.26 & -- \\
M3T \cite{14}       & 0.7895 & 0.7633 & 0.7549 & 0.7516 & 0.6768 & \textcolor{red}{0.9253} & 153.20 & 29.23 & -- \\
AAGN \cite{15}      & 0.7281 & 0.8369 & 0.6771 & 0.5814 & 0.5811 & \textcolor{blue}{0.9021} & 589.50 & 197.51 & -- \\
ResSNN \cite{36}    & 0.6372 & 0.5411 & 0.6372 & 0.5649 & 0.4299 & 0.8430 & 1.78  & 5.85 & 4 \\
VGGSNN \cite{31}    & 0.6814 & 0.5247 & 0.6814 & 0.5921 & 0.4882 & 0.8773 & 1.62  & 10.08 & 16 \\
TCJA \cite{32}      & 0.5752 & 0.5701 & 0.5752 & 0.4901 & 0.3598 & 0.8312 & 3.95  & \textcolor{red}{2.87} & 16 \\
SPT \cite{35}       & 0.6930 & 0.6651 & 0.6415 & 0.6127 & 0.5202 & 0.8410 & \textcolor{blue}{1.55} & 9.64 & 2 \\
PointNet \cite{34}  & 0.5579 & 0.5569 & 0.5658 & 0.5505 & 0.3394 & 0.6983 & \textcolor{red}{0.73} & \textcolor{blue}{3.47} & 16 \\
Ours                & \textcolor{red}{0.8155} & 0.8123 & \textcolor{red}{0.8122} & \textcolor{red}{0.8122} & \textcolor{red}{0.7214} & 0.8730 & 1.88  & 43.11 & 2 \\
\hline
\end{tabularx}
\end{table}
\normalcolor  

\textcolor{black}{%
As shown in Fig.~\ref{rose} and Fig.~\ref{rodar},our proposed model demonstrates superior overall performance compared to existing state-of-the-art models. The experimental results detailed in Table~\ref{tab:ADNI_five_folds}  and ~\ref{tab:AIBL_five_folds} indicate that, with only two time steps, the model achieved an average Classification Accuracy of $0.8325$ and an average AUC of $0.9294$ across five-fold cross-validation on the ADNI dataset. Furthermore, the model attained a Kappa coefficient of $0.7408$ on the dedicated test set, signifying strong reliability in its predictions. Beyond classification performance, the model also exhibits exceptional computational efficiency, as shown in the result of Table~\ref{tab:ADNI_test} and Table~\ref{tab:AIBL_test}, achieving a speedy training time of only $20 \text{ s}$ and an energy consumption of merely $1.74 \text{ J}$.}

\textcolor{black}{%
Compared with ResNet-based and fully convolutional network models, our proposed model demonstrated consistently superior performance. The average Classification Accuracy showed an improvement ranging from 1.53\% to 2.65\% throughout the five-fold cross-validation process. In addition, the average AUC was significantly enhanced, reaching 0.9294, whereas the ResNet architectures and FCN models achieved values of only 0.9064, 0.9152, and 0.9051, respectively. This performance advantage was also consistently observed on the independent test set.While our model incorporates residual connections to address the vanishing gradient issue in deep networks, a mechanism also utilized in ResNet architectures, it differentiates itself through the integration of LIF neurons, which encourage feature sparsity, combined with the distinctive convolutional structure of the FasterSNNBlock. Together, these elements contributed to exceptionally low energy consumption, measured at only 1.74 J, which is markedly lower than the 3.68 J and 9.54 J consumed by the ResNet models, as well as the 14.77 J required by the FCN model. These results collectively underscore the dual advantage of our model in terms of both energy efficiency and performance, supporting its potential suitability for clinical applications.
}

\textcolor{black}{%
Transformer-based models including ViT and M3T employ self-attention mechanisms to improve the model's ability to focus on pathological regions. Nevertheless, these mechanisms come with considerable computational overhead, especially in the context of 3D volumetric data processing. The energy consumption of ViT and M3T, which respectively amount to 7.14 J and 148.64 J, is substantially higher than the 1.74 J consumed by our proposed model. Although our model contains 43.11 million parameters, it accomplishes a single training epoch in merely 20 seconds, a duration significantly shorter than the 106 seconds required by M3T. In terms of evaluation metrics, the proposed model attained a classification accuracy of 0.8310 and a Kappa coefficient of 0.7408 on the test set, thereby consistently exceeding the corresponding results of ViT, with an accuracy of 0.7257 and a Kappa value of 0.5518, as well as those of M3T, which achieved an accuracy of 0.8246 and a Kappa value of 0.7287. Together, these quantitative findings highlight the proposed model's distinct advantages in classification performance, predictive consistency, and overall computational efficiency.
}

\textcolor{black}{%
Compared to the AAGN model, the proposed model offers significant and consistent advantages across energy consumption, training time, and classification metrics. The AAGN model requires an exceedingly high energy expenditure of $622.29 \text{ J}$ and takes $287 \text{ s}$ to complete a single training epoch. In contrast, our proposed model consumes only $1.74 \text{ J}$ and finishes training in just $20 \text{ s}$. Furthermore, the proposed model demonstrated significantly superior performance over AAGN, achieving an average Classification Accuracy of $0.8325$ and an average AUC of $0.9294$ in the five-fold cross-validation, compared to AAGN's corresponding values of $0.7080$ and $0.8602$.
}

\textcolor{black}{%
Compared to other existing spiking neural network models, our proposed architecture demonstrated superior performance across multiple key metrics. In terms of energy consumption, our model consumed 1.74 J, which is comparable to the 1.64 J of ResSNN and the 1.76 J of VGGSNN, yet lower than the 3.61 J required by TCJA, indicating a favorable level of feature sparsity in our network. Regarding classification accuracy, the proposed model achieved a value of 0.8310 on the test set, significantly surpassing ResSNN at 0.6903, PointNet at 0.7579, VGGSNN at 0.6930, TCJA at 0.7257, and SPT at 0.7699. Furthermore, the average AUC of our model reached 0.9406, exceeding that of all competing SNN models and definitively establishing its dominance in classification performance.}

\textcolor{black}{%
Furthermore, as evident from Table~\ref{tab:ADNI_five_folds} and Table~\ref{tab:AIBL_five_folds}, most existing SNN models exhibit considerable instability accompanied by significant performance fluctuations. Among these models, PointNet demonstrates the most severe fluctuation, reflected in an accuracy of 0.5907 with a standard deviation of 0.1692. In contrast, SPT shows minimal fluctuation, as indicated by its accuracy of 0.7567 and a standard deviation of 0.0124, yet it achieves only relatively low overall accuracy. Our proposed model strikes an excellent and robust balance between diagnostic performance and stability, achieving an accuracy of 0.8325 with a low standard deviation of 0.0276. To further rigorously validate the model's superiority, we systematically compared FasterSNN against the best suboptimal model on the independent AIBL dataset. Using a paired bootstrap statistical test with 5,000 iterations, we obtained a p-value of 0.0344, which falls below the conventional significance threshold of 0.05. This statistically significant outcome convincingly confirms that FasterSNN delivers consistently reliable and superior performance compared to the suboptimal baseline on the AIBL dataset.}

\subsubsection{Visualization and Stability Assessment}
\begin{figure}[H]
\centering
\includegraphics[width=\textwidth]{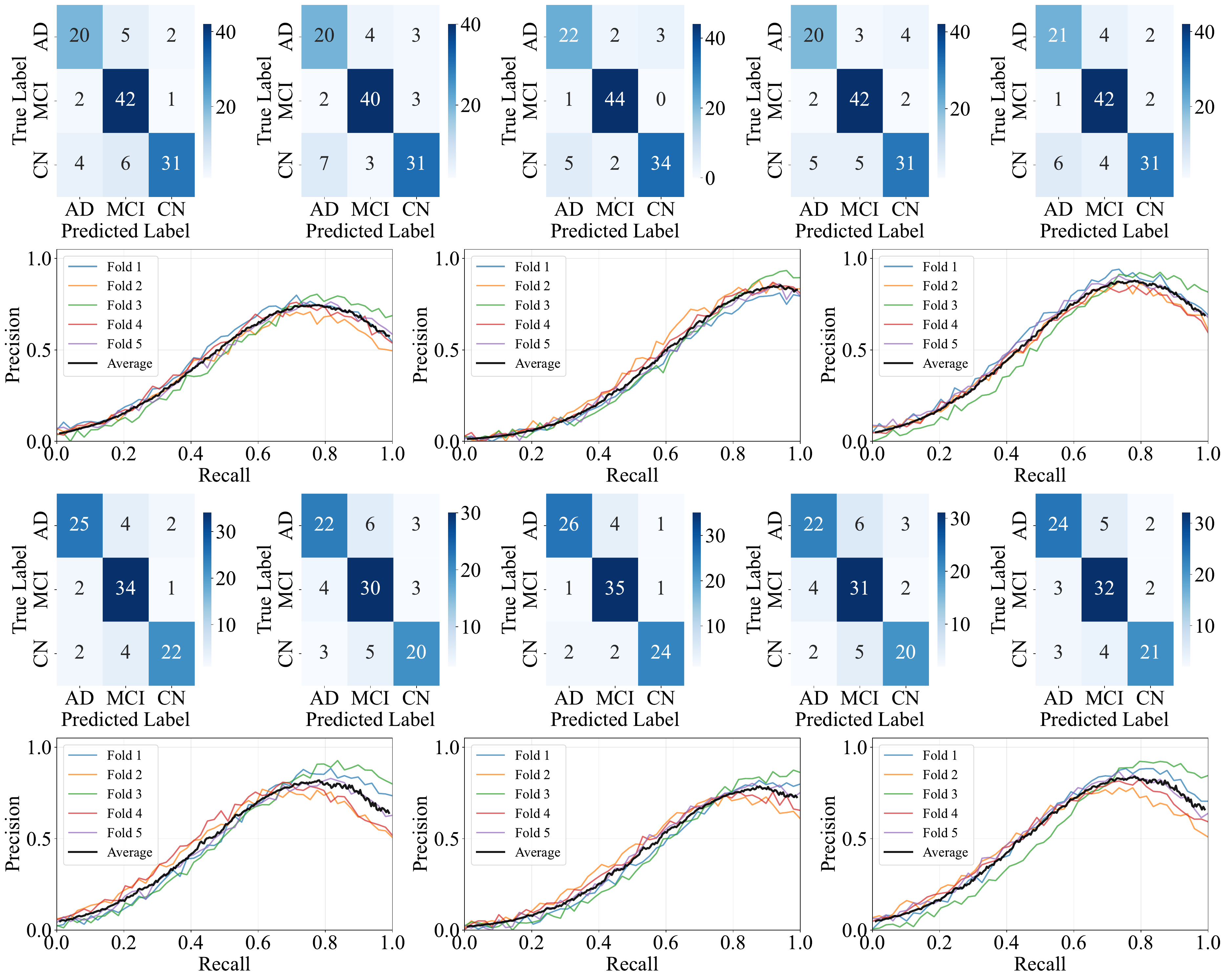}
\caption{\color{black}Model performance evaluation across ADNI and AIBL datasets. Rows 1 and 3 display the confusion matrices for each fold of the 5-fold cross-validation on the ADNI and AIBL datasets, respectively. Rows 2 and 4 present the precision-recall curves for the three diagnostic classes on the ADNI and AIBL datasets correspondingly.}
\label{cm_precision}
\end{figure}

\textcolor{black}{%
To rigorously assess the model's stability during training, we recorded the confusion matrices and class-specific Precision-Recall (PR) curves on the validation set for each of the five folds in the cross-validation procedure.\\
As clearly illustrated in Fig.~\ref{cm_precision}, the confusion matrices and the class-wise Precision-Recall curves, the model exhibited remarkably similar and consistent classification performance across all five folds. The consistently high overall consistency across the folds decisively affirms the inherent stability and generalizability of our model. Furthermore, the PR curves for all diagnostic classes showed a highly overlapping trend across the five folds, with only minor and negligible fluctuations, further corroborating the robust robustness and high reliability of the model training process.
}

\textcolor{black}{%
Furthermore, to provide a more intuitive and direct evaluation of our model's capability in the complex multi-class classification task, we generated and presented confusion matrices detailing the classification outcomes on the fully independent ADNI and AIBL test sets. These matrices enable a critical and comprehensive examination of the model's ability to discriminate among various clinical disease stages and to assess its strong generalization capacity.}

\begin{figure}[H]
\centering  
\includegraphics[width=\linewidth]{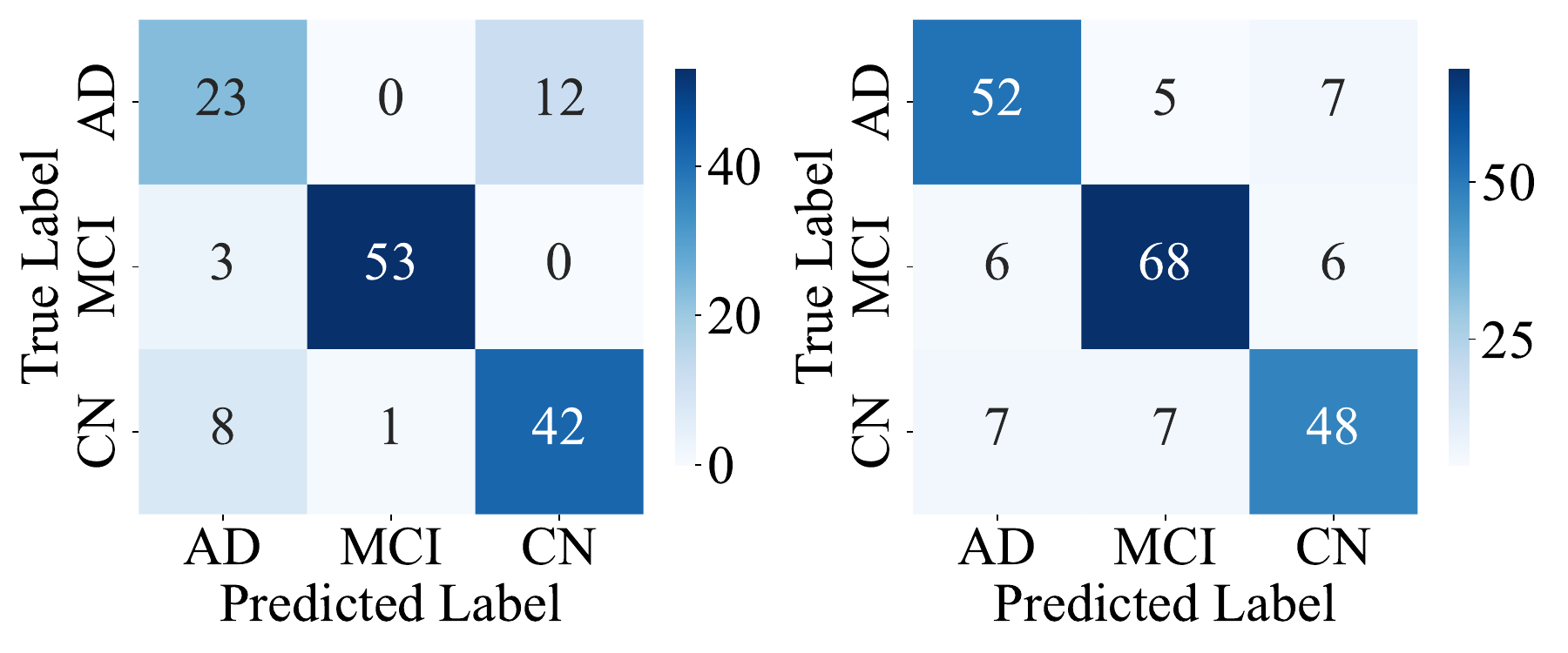} 
\caption{\color{black}Confusion matrices of FasterSNN on the ADNI (left) and AIBL (right) datasets, illustrating model's performance across different classes.}
\label{cm}
\end{figure}

\textcolor{black}{%
As demonstrated in Fig.~\ref{cm},the 56 samples belonging to the MCI class, the model accurately classified 53 samples. This result highlights the model's robust ability to detect MCI, a critical stage for therapeutic intervention and early screening of AD. For the AD class, only 23 out of 29 samples were correctly identified. The CN class achieved an accuracy of $82.3\%$, correctly classifying 42 samples; however, eight samples were erroneously classified as AD, and 1 sample was incorrectly classified as MCI.
}

\textcolor{black}{%
Analyzing the diagnostic performance specifically on the AIBL dataset reveals the following: For the AD class, a total of 64 samples were present. The model correctly identified 52 of these, with only five samples being misclassified as MCI and seven samples as CN, resulting in a classification accuracy of $81.25\%$. The model exhibited equally strong performance in the MCI class, achieving a high correct classification rate of 
$85\%$. Out of 80 samples, 68 were accurately identified, despite six being misclassified as AD and another six as CN. In the CN class, 48 out of 62 samples were correctly identified, while seven were misclassified as MCI and seven as AD.
}

\textcolor{black}{%
Crucially, the model consistently achieved outstanding classification performance for the MCI category across both the ADNI and AIBL datasets. This result underscores the extremely high reliability of our model in identifying MCI, which represents the pivotal stage for therapeutic intervention and effective early screening in the continuum of Alzheimer's Disease progression.
}

\subsection{Ablation Experiments}
\textcolor{black}{%
To validate the necessity and efficacy of our designed modules and selected hyperparameters, we conducted a series of ablation studies. Specifically, these experiments investigated the individual contributions of the LIF and MSF modules, as well as the influence of key parameters, such as the time step and input image resolution, on overall model performance.
}\\

\begin{table}[H]
    \centering
    \renewcommand{\tabularxcolumn}[1]{>{\centering\arraybackslash}m{#1}}
    \tiny
    \renewcommand{\arraystretch}{1.2}
    \caption{\color{black}Performance Comparison of Different Configurations on the ADNI dataset.Red and blue denote best and second-best performance respectively.}
    \label{tab:ablation_adni}
    \color{black}
    \setlength{\arrayrulewidth}{0.5pt}
    \begin{tabularx}{\textwidth}{@{}l*{8}{X}@{}}
        \hline
        & \multicolumn{6}{c}{Metrics ($\uparrow$)} & \multicolumn{2}{c}{Resources ($\downarrow$)} \\
        \cline{2-9}
        Configuration & Accuracy & Precision & Recall & F1-score & Kappa & AUC & Energy (J) & Train Time (s) \\
        \hline
        w/o LIF        & 0.8010                     & 0.7974                     & 0.7942                     & 0.7953                     & 0.6986                     & 0.8181                     & 2.85              & 22 \\
        w/o MSF        & 0.7621                     & 0.7522                     & 0.7500                     & 0.7492                     & 0.6391                     & 0.8794                     & \textcolor{blue}{1.74} & \textcolor{blue}{20} \\
        Resolution 96  & \textcolor{red}{0.8369}    & 0.8094                     & 0.8144                     & 0.8113                     & 0.7215                     & \textcolor{red}{0.9426}    & 6.34              & 74 \\
        Resolution 128 & 0.8239                     & 0.8087                     & 0.8042                     & 0.8059                     & 0.7300                     & 0.9314                     & 14.08             & 187 \\
        Time Step 1    & 0.7972                     & 0.8196                     & 0.7972                     & 0.7715                     & 0.6807                     & 0.9023                     & \textcolor{red}{0.89} & \textcolor{red}{11} \\
        Time Step 4    & \textcolor{blue}{0.8343}   & \textcolor{red}{0.8363}    & \textcolor{red}{0.8343}    & \textcolor{blue}{0.8306}   & \textcolor{red}{0.7449}    & 0.9354                     & 9.05              & 72 \\
        Ours           & 0.8310                     & \textcolor{blue}{0.8346}   & \textcolor{blue}{0.8310}   & \textcolor{red}{0.8312}    & \textcolor{blue}{0.7408}   & \textcolor{blue}{0.9406}   & \textcolor{blue}{1.74} & \textcolor{blue}{20} \\
        \hline
    \end{tabularx}
\end{table}

\begin{table}[H]
    \centering
    \renewcommand{\tabularxcolumn}[1]{>{\centering\arraybackslash}m{#1}}
    \tiny
    \renewcommand{\arraystretch}{1.2}
    \caption{\color{black}Performance Comparison of Different Configurations on the AIBL dataset.Red and blue denote best and second-best performance respectively.}
    \label{tab:ablation_aibl}
    \color{black}
    \setlength{\arrayrulewidth}{0.5pt}
    \begin{tabularx}{\textwidth}{@{}l*{8}{X}@{}}
        \hline
        & \multicolumn{6}{c}{Metrics ($\uparrow$)} & \multicolumn{2}{c}{Resources ($\downarrow$)} \\
        \cline{2-9}
        Configuration & Accuracy & Precision & Recall & F1-score & Kappa & AUC & Energy (J) & Train Time (s) \\
        \hline
        w/o LIF        & 0.8010                     & 0.7974                     & 0.7942                     & 0.7953                     & 0.6986                     & 0.8547                     & 3.12              & 20 \\
        w/o MSF        & 0.7621                     & 0.7522                     & 0.7500                     & 0.7492                     & 0.6391                     & 0.8337                     & \textcolor{blue}{1.74} & \textcolor{blue}{16} \\
        Resolution 96  & \textcolor{blue}{0.8204}   & \textcolor{blue}{0.8148}   & \textcolor{blue}{0.8200}   & \textcolor{red}{0.8182}    & \textcolor{red}{0.7298}    & 0.8706                     & 6.68              & 65 \\
        Resolution 128 & \textcolor{red}{0.8252}    & \textcolor{red}{0.8288}    & \textcolor{red}{0.8277}    & \textcolor{blue}{0.8174}   & 0.7161                     & \textcolor{red}{0.9263}    & 15.42             & 158 \\
        Time Step 1    & 0.8025                     & 0.7987                     & 0.8025                     & 0.7972                     & 0.6950                     & \textcolor{blue}{0.9229}   & \textcolor{red}{1.02} & \textcolor{red}{11} \\
        Time Step 4    & 0.8148                     & 0.8134                     & 0.8148                     & 0.8107                     & 0.7133                     & 0.9166                     & 9.78              & 68 \\
        Ours           & 0.8155                     & 0.8123                     & 0.8122                     & 0.8122                     & \textcolor{blue}{0.7214}   & 0.8730                     & 1.88              & 17 \\
        \hline
    \end{tabularx}
\end{table}

\begin{figure}[h]
\centering
\includegraphics[width=\linewidth]{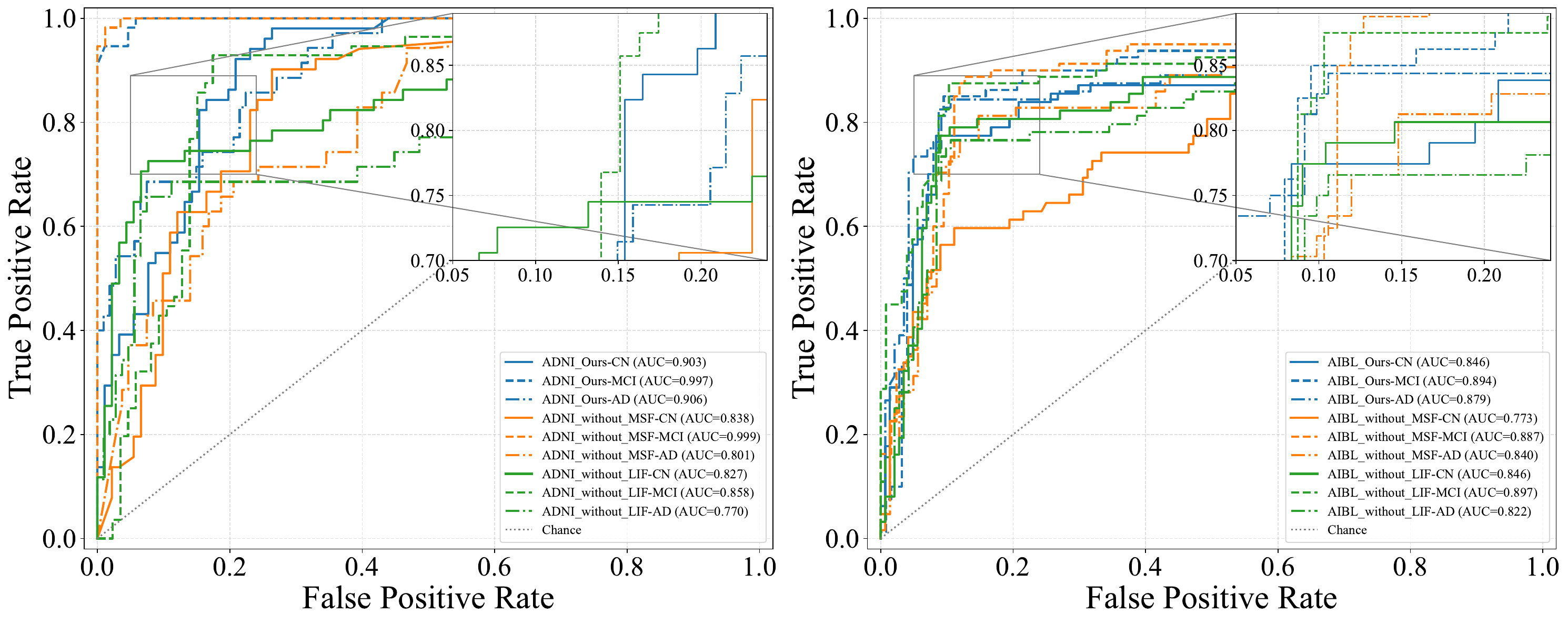}
\caption{\color{black}ROC curves of model ablation experiments on the ADNI (left) and AIBL (right) datasets. The x-axis denotes the false positive rate (FPR), and the y-axis denotes the true positive rate (TPR). Each curve represents a different model configuration, with corresponding AUC values indicated. To highlight performance differences, the upper-left region of each plot is zoomed in.}
\label{roc}
\end{figure}
\vspace{-1.0em}

\begin{figure}[h]
\centering
\includegraphics[width=\linewidth]{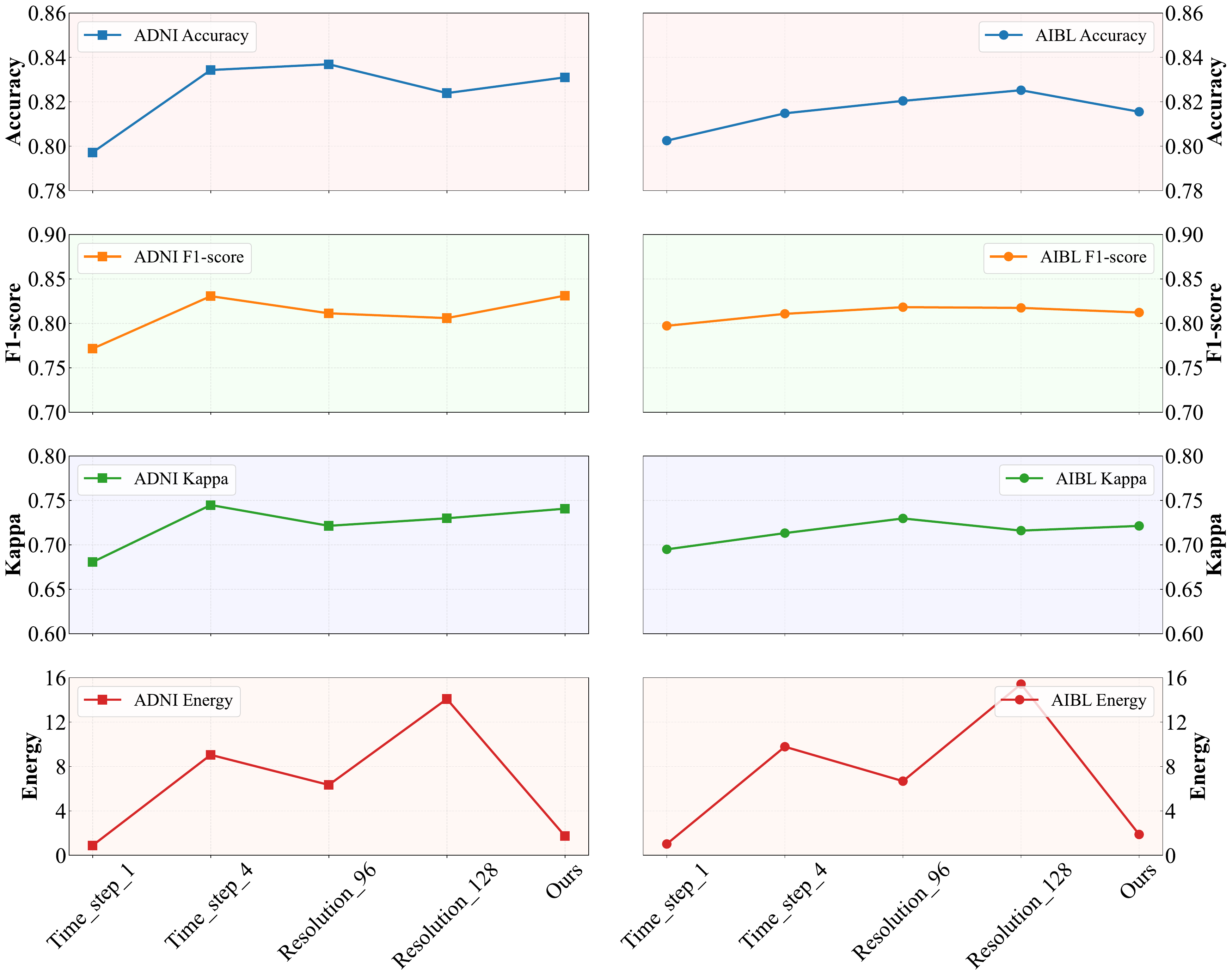}
\caption{\color{black}Comparative evaluation of model configurations on the ADNI (left) and AIBL (right) datasets. Each subplot column corresponds to a different model configuration: Time Step 1, Time Step 4, Resolution 96, Resolution 128, and the proposed method (Ours). From top to bottom, the rows display the performance metrics of Accuracy, F1-score, Cohen's Kappa, and Energy.}
\label{zhexian}
\end{figure}

\textcolor{black}{%
Tables \ref{tab:ablation_adni} and \ref{tab:ablation_aibl} present the ablation experimental results of the model on the ADNI and AIBL datasets, evaluating key components, input resolution, and temporal steps to assess the contribution of each module and the rationale behind the default parameter settings.
}

\textcolor{black}{%
As shown in Fig.~\ref{roc}, in the key component ablation experiments, removing the LIF neurons significantly impaired model performance. Compared to the complete model, Accuracy, Precision, Recall, and F1-score decreased by 3.00\%, 3.72\%, 3.68\%, and 3.59\%, respectively. The Kappa coefficient dropped from 0.7408 to 0.6986, and the AUC decreased from 0.9406 to 0.8181. Meanwhile, energy consumption increased from 1.74 J to 2.85 J, and training time rose from 20 s to 22 s. This indicates that LIF neurons effectively reduce computational overhead through sparse spike activation while enhancing model consistency and discriminative capability.
}

\textcolor{black}{%
In contrast, removing the MSF module had a more substantial impact on performance. Accuracy, Precision, Recall, and F1-score decreased by 6.89\%, 8.24\%, 8.10\%, and 8.20\%, respectively, with Kappa dropping to 0.6391 and AUC to 0.8794. Notably, the removal of the MSF module had almost no effect on energy consumption and training time, indicating its low computational overhead and significant performance contribution. The results demonstrate that both LIF neurons and the MSF module are indispensable key components of the model, working synergistically to ensure high accuracy while achieving energy efficiency optimization.
}

\textcolor{black}{%
As shown in Fig.~\ref{zhexian}, in the ablation study on input resolution, increasing the resolution yielded only a limited performance improvement at the expense of a substantial increase in computational cost. For instance, when the resolution was escalated from $64^3$ to $96^3$ and $128^3$, the Classification Accuracy only increased by $0.49\%$ and $0.97\%$, respectively. However, energy consumption consequently rose significantly to $6.68$ J and $15.42$ J, and the training time also increased dramatically from $17$ s to $65$ s and $158$ s. Given the primary objectives of achieving lightweight design and high efficiency for potential clinical translation, the $64^3$ resolution was selected as the default setting. This choice represents the optimal balance between diagnostic performance and computational overhead.}

\textcolor{black}{%
In the temporal step-size ablation study, increasing the temporal step size generally improved temporal feature modeling. Specifically, when the temporal step size was increased from $T=1$ to $T=2$, the Accuracy improved from $0.7972$ to $0.8310$, the Kappa coefficient increased from $0.6807$ to $0.7408$, and the AUC rose from $0.9023$ to $0.9406$. This improvement was accompanied by only a moderate increase in energy consumption to $1.74\ J$ and an extension of training time to $20\ s$. However, further increasing the temporal steps to $T=4$ resulted in only marginal performance gains: Accuracy increased by merely $0.33\%$, Kappa improved by $0.41\%$, and the AUC even slightly decreased to $0.9354$. Concurrently, both the energy consumption and the training time surged significantly to $9.05\ J$ and $72\ s$, respectively. Consequently, the results clearly indicate that $T=2$ achieves the optimal trade-off between performance enhancement and computational efficiency. Thus, $T=2$ was selected as the default temporal step size setting.
}

\subsection{Interpretability Analysis}

 \begin{figure}[H]
\centering
\includegraphics[width=\textwidth]{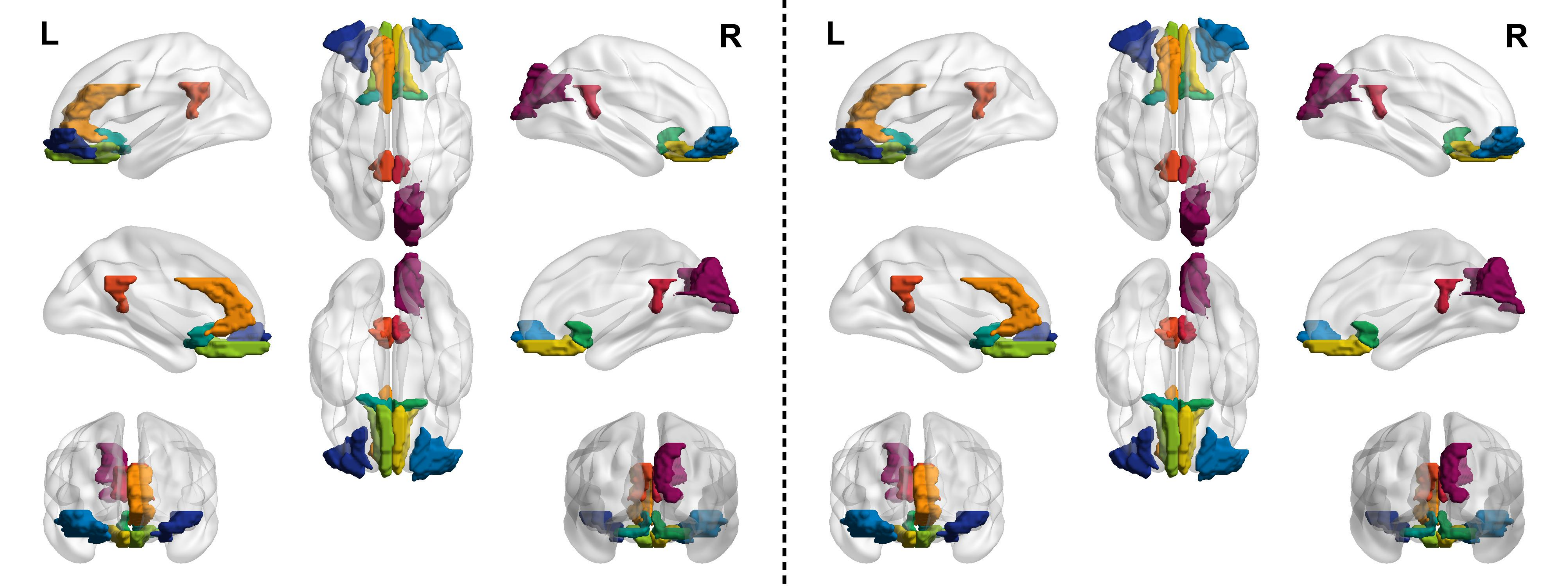}
\caption{\color{black}Visualization of spatial attention projected onto skull-stripped brain 3D MR images. The left panel shows the attention maps derived from the ADNI dataset, while the right panel presents corresponding visualizations from the AIBL dataset.}
\label{ADNNI_tif}
\end{figure}

\textcolor{black}{%
Regarding the model interpretability, we mapped the attention maps generated by the SWA module onto the three-dimensional skull-stripped brain structure, as shown in Fig.~\ref{ADNNI_tif}. The visualization results indicate that the model's high-response regions are predominantly concentrated in the entorhinal cortex, parahippocampal gyrus, posterior cingulate cortex, and dorsolateral prefrontal cortex. This spatial distribution exhibits a high degree of consistency with the established key pathological sites recognized in early AD. Although a rigorous quantitative analysis of spatial overlap has not yet been conducted, this finding is anatomically consistent with previous neuroimaging research discoveries, thereby providing preliminary support for the plausibility of the model's explanation. Notably, the model's significant focus on the entorhinal cortex further strengthens the biological basis for its accurate discrimination of the MCI class.
}

\subsection{Limitations Analysis}
\textcolor{black}{%
Although FasterSNN demonstrates excellent performance and practicality across multiple dimensions, this study has several noteworthy limitations:}

\textcolor{black}{%
During the data preprocessing phase, this study did not incorporate advanced harmonization techniques, such as ComBat correction, non-linear MNI registration, or skull stripping, to systematically adjust for cross-center and cross-device image variations. Although we mitigated the impact of data source heterogeneity through basic measures like Z-score standardization and spatial resampling, potential discrepancies in acquisition protocols, scanning parameters, and equipment models persist between the ADNI and AIBL datasets. Consequently, these residual heterogeneities may introduce subtle but non-negligible biases in the input images. Future work will focus on integrating more sophisticated preprocessing methods to perform finer-grained debiasing of the MRI scans, thereby further enhancing image quality and significantly improving the robustness of the diagnostic model.
}

\textcolor{black}{%
The model currently relies solely on single-modality MRI data for both training and inference, without incorporating complementary neuroimaging modalities. While this single-modality approach offers clear advantages in terms of reduced deployment costs and simplified system workflows, it also inherently limits the model's capacity to capture complex pathological mechanisms. Subsequent research will explore the integration of multimodal information, such as PET imaging or diffusion tensor imaging, to enhance the model's clinical applicability and diagnostic comprehensiveness.
}

\textcolor{black}{%
The interpretability analysis of the model remains primarily qualitative and visualization-based, lacking a quantitative alignment mechanism with anatomical priors or expert annotations. Although the attention maps generated by the SWA module show visual correspondence with key AD-related regions such as the hippocampus and entorhinal cortex, the absence of quantitative metrics undermines their persuasiveness in clinical settings. Future work will introduce brain atlas templates or atlas-based ROI annotations to conduct spatial consistency evaluations, thereby establishing a more rigorous and verifiable interpretability framework.
}

\textcolor{black}{%
The training and validation of the model were predominantly based on the ADNI and AIBL datasets, with generalization to other independent cohorts yet to be thoroughly investigated. To enhance the model's fairness and generalizability, subsequent studies will incorporate additional datasets, such as OASIS, and employ cross-population transfer learning and federated training strategies to systematically evaluate the model's stability and robustness across diverse demographic and clinical settings.
}

\section{Conclusion}
This study introduces FasterSNN, a biologically inspired hybrid neural architecture that combines LIF neurons, region-adaptive convolutions, and a multi-scale spiking attention mechanism. The proposed design effectively addresses key challenges in AD diagnosis, such as high energy consumption and limited interpretability found in traditional deep learning approaches. By utilizing the sparse, event-driven computations of SNNs alongside hierarchical multi-scale feature fusion, FasterSNN delivers remarkably robust classification performance while significantly lowering computational costs. Experimental results on the ADNI and AIBL datasets demonstrate the clear superiority of FasterSNN over state-of-the-art CNNs, transformers, and other SNN models. FasterSNN achieves consistently higher diagnostic accuracy and efficiency across multiple evaluation metrics. Ablation studies further highlight the critical contributions of the LIF neurons and the MSF module, which together substantially enhance the model's reliability and energy efficiency. Despite these advancements, the current model relies solely on unimodal MRI data and operates with fixed time steps, which may restrict its adaptability to more complex and diverse clinical scenarios. Future research will focus on incorporating dynamic temporal modeling and exploring multimodal data fusion strategies to improve overall generalizability and broaden the model's clinical applicability.

\section*{CRediT authorship contribution statement}
\textbf{Changwei Wu}: Methodology, Visualization, Writing – original draft.
\textbf{Yifei Chen}: Conceptualization, Methodology, Writing – original draft.
\textbf{Yuxin Du}: Validation, Visualization.
\textbf{Jinying Zong}: Validation, Visualization.
\textbf{Jie Dong}: Validation, Visualization.
\textbf{Mingxuan Liu}: Validation, Visualization.
\textbf{Yong Peng}: Writing – review \& editing.
\textbf{Jin Fan}: Writing – review \& editing.
\textbf{Feiwei Qin}: Project administration, Writing – review \& editing.
\textbf{Chaomiao Wang}: Supervision, Writing – review \& editing.

\section*{Declaration of competing interest}
The authors declare that they have no known competing financial interests or personal relationships that could have appeared to influence the work reported in this paper.

\section*{Acknowledgments}
This work was supported by the Fundamental Research Funds for the Provincial Universities of Zhejiang (No. GK259909299001-006), Anhui Provincial Joint Construction Key Laboratory of Intelligent Education Equipment and Technology (No. IEET202401), Guangdong Basic and Applied Basic Research Foundation (No. 2025A1515011617, 2022A1515110570) and Innovation Teams of Youth Innovation in Science and Technology of High Education Institutions of Shandong Province (No. 2021KJ088).


\bibliographystyle{elsarticle-num} 

\begin{thebibliography}{10}
\providecommand{\url}[1]{#1}
\csname url@samestyle\endcsname
\providecommand{\newblock}{\relax}
\providecommand{\bibinfo}[2]{#2}
\providecommand{\BIBentrySTDinterwordspacing}{\spaceskip=0pt\relax}
\providecommand{\BIBentryALTinterwordstretchfactor}{4}
\providecommand{\BIBentryALTinterwordspacing}{\spaceskip=\fontdimen2\font plus
\BIBentryALTinterwordstretchfactor\fontdimen3\font minus \fontdimen4\font\relax}
\providecommand{\BIBforeignlanguage}[2]{{%
\expandafter\ifx\csname l@#1\endcsname\relax
\typeout{** WARNING: IEEEtran.bst: No hyphenation pattern has been}%
\typeout{** loaded for the language `#1'. Using the pattern for}%
\typeout{** the default language instead.}%
\else
\language=\csname l@#1\endcsname
\fi
#2}}
\providecommand{\BIBdecl}{\relax}
\BIBdecl

\bibitem{1}
C.~R. Jack~Jr, D.~A. Bennett, K.~Blennow, M.~C. Carrillo, B.~Dunn, S.~B. Haeberlein, D.~M. Holtzman, W.~Jagust, F.~Jessen, J.~Karlawish \emph{et~al.}, ``Nia-aa research framework: Toward a biological definition of alzheimer's disease,'' \emph{Alzheimer's \& Dementia}, vol.~14, no.~4, pp. 535--562, 2018.

\bibitem{2}
T.~G. Beach, S.~E. Monsell, L.~E. Phillips, and W.~Kukull, ``Accuracy of the clinical diagnosis of alzheimer disease at national institute on aging alzheimer disease centers, 2005--2010,'' \emph{Journal of Neuropathology and Experimental Neurology}, vol.~71, no.~4, pp. 266--273, 2012.

\bibitem{3}
J.~Chen, S.-h. Kao, H.~He, W.~Zhuo, S.~Wen, C.-H. Lee, and S.-H.~G. Chan, ``Run, don't walk: Chasing higher flops for faster neural networks,'' in \emph{Proceedings of the IEEE/CVF Conference on Computer Vision and Pattern Recognition}, 2023, pp. 12\,021--12\,031.

\bibitem{10}
X.-a. Bi, Q.~Shu, Q.~Sun, and Q.~Xu, ``Random support vector machine cluster analysis of resting-state fmri in alzheimer's disease,'' \emph{PLOS One}, vol.~13, no.~3, p. e0194479, 2018.

\bibitem{11}
S.~Qiu, P.~S. Joshi, M.~I. Miller, C.~Xue, X.~Zhou, C.~Karjadi, G.~H. Chang, A.~S. Joshi, B.~Dwyer, S.~Zhu \emph{et~al.}, ``Development and validation of an interpretable deep learning framework for alzheimer’s disease classification,'' \emph{Brain}, vol. 143, no.~6, pp. 1920--1933, 2020.

\bibitem{12}
R.~Kushol, A.~Masoumzadeh, D.~Huo, S.~Kalra, and Y.-H. Yang, ``Addformer: Alzheimer’s disease detection from structural mri using fusion transformer,'' in \emph{2022 IEEE 19th International Symposium on Biomedical Imaging (ISBI)}.\hskip 1em plus 0.5em minus 0.4em\relax IEEE, 2022, pp. 1--5.

\bibitem{14}
J.~Jang and D.~Hwang, ``M3t:three-dimensional medical image classifier using multi-plane and multi-slice transformer,'' in \emph{Proceedings of the IEEE/CVF Conference on Computer Vision and Pattern Recognition}, 2022, pp. 20\,718--20\,729.

\bibitem{15}
H.~Jiang and C.~Miao, ``Anatomy-aware gating network for explainable alzheimer’s disease diagnosis,'' in \emph{International Conference on Medical Image Computing and Computer-Assisted Intervention}.\hskip 1em plus 0.5em minus 0.4em\relax Springer, 2024, pp. 90--100.

\bibitem{16}
M.~Liu, J.~Tang, Y.~Chen, H.~Li, J.~Qi, S.~Li, K.~Wang, J.~Gan, Y.~Wang, and H.~Chen, ``Spiking-physformer: Camera-based remote photoplethysmography with parallel spike-driven transformer,'' \emph{Neural Networks}, vol. 185, p. 107128, 2025.

\bibitem{17}
X.~Lin, M.~Liu, K.~Liu, and H.~Chen, ``Spike-slr: an energy-efficient parallel spiking transformer for event-based sign language recognition,'' 2024.

\bibitem{20}
B.~Magnin, L.~Mesrob, S.~Kinkingn{\'e}hun, M.~P{\'e}l{\'e}grini-Issac, O.~Colliot, M.~Sarazin, B.~Dubois, S.~Leh{\'e}ricy, and H.~Benali, ``Support vector machine-based classification of alzheimer’s disease from whole-brain anatomical mri,'' \emph{Neuroradiology}, vol.~51, pp. 73--83, 2009.

\bibitem{21}
A.~Ortiz, J.~M. G{\'o}rriz, J.~Ram{\'\i}rez, F.~J. Mart{\'\i}nez-Murcia, A.~D.~N. Initiative \emph{et~al.}, ``Lvq-svm based cad tool applied to structural mri for the diagnosis of the alzheimer’s disease,'' \emph{Pattern Recognition Letters}, vol.~34, no.~14, pp. 1725--1733, 2013.

\bibitem{22}
A.~Retico, P.~Bosco, P.~Cerello, E.~Fiorina, A.~Chincarini, and M.~E. Fantacci, ``Predictive models based on support vector machines: Whole-brain versus regional analysis of structural mri in the alzheimer's disease,'' \emph{Journal of Neuroimaging}, vol.~25, no.~4, pp. 552--563, 2015.

\bibitem{23}
Y.~Vichianin, A.~Khummongkol, P.~Chiewvit, A.~Raksthaput, S.~Chaichanettee, N.~Aoonkaew, and V.~Senanarong, ``Accuracy of support-vector machines for diagnosis of alzheimer's disease, using volume of brain obtained by structural mri at siriraj hospital,'' \emph{Frontiers in Neurology}, vol.~12, p. 640696, 2021.

\bibitem{24}
R.~E. Turkson, H.~Qu, C.~B. Mawuli, and M.~J. Eghan, ``Classification of alzheimer’s disease using deep convolutional spiking neural network,'' \emph{Neural Processing Letters}, vol.~53, no.~4, pp. 2649--2663, 2021.

\bibitem{25}
A.~Lebedev, E.~Westman, G.~Van~Westen, M.~Kramberger, A.~Lundervold, D.~Aarsland, H.~Soininen, I.~K{\l}oszewska, P.~Mecocci, M.~Tsolaki \emph{et~al.}, ``Random forest ensembles for detection and prediction of alzheimer's disease with a good between-cohort robustness,'' \emph{NeuroImage: Clinical}, vol.~6, pp. 115--125, 2014.

\bibitem{26}
Y.~Wu, L.~Deng, G.~Li, J.~Zhu, and L.~Shi, ``Spatio-temporal backpropagation for training high-performance spiking neural networks,'' \emph{Frontiers in Neuroscience}, vol.~12, p. 331, 2018.

\bibitem{30}
K.~He, X.~Zhang, S.~Ren, and J.~Sun, ``Deep residual learning for image recognition,'' in \emph{Proceedings of the IEEE Conference on Computer Vision and Pattern Recognition}, 2016, pp. 770--778.

\bibitem{31}
S.~Gou, J.~Fu, Y.~Sha, Z.~Cao, Z.~Guo, J.~K. Eshraghian, R.~Li, and L.~Jiao, ``Dynamic spatio-temporal pruning for efficient spiking neural networks,'' \emph{Frontiers in Neuroscience}, vol.~19, p. 1545583, 2025.

\bibitem{32}
R.-J. Zhu, M.~Zhang, Q.~Zhao, H.~Deng, Y.~Duan, and L.-J. Deng, ``Tcja-snn: Temporal-channel joint attention for spiking neural networks,'' \emph{IEEE Transactions on Neural Networks and Learning Systems}, 2024.

\bibitem{33}
K.~Kunanbayev, V.~Shen, and D.-S. Kim, ``Training vit with limited data for alzheimer’s disease classification: An empirical study,'' in \emph{International Conference on Medical Image Computing and Computer-Assisted Intervention}.\hskip 1em plus 0.5em minus 0.4em\relax Springer, 2024, pp. 334--343.

\bibitem{34}
D.~Ren, Z.~Ma, Y.~Chen, W.~Peng, X.~Liu, Y.~Zhang, and Y.~Guo, ``Spiking pointnet: Spiking neural networks for point clouds,'' \emph{Advances in Neural Information Processing Systems}, vol.~36, pp. 41\,797--41\,808, 2023.

\bibitem{35}
P.~Wu, B.~Chai, H.~Li, M.~Zheng, Y.~Peng, Z.~Wang, X.~Nie, Y.~Zhang, and X.~Sun, ``Spiking point transformer for point cloud classification,'' in \emph{Proceedings of the AAAI Conference on Artificial Intelligence}, vol.~39, no.~20, 2025, pp. 21\,563--21\,571.

\bibitem{36}
Y.~Hu, L.~Deng, Y.~Wu, M.~Yao, and G.~Li, ``Advancing spiking neural networks toward deep residual learning,'' \emph{IEEE Transactions on Neural Networks and Learning Systems}, vol.~36, no.~2, pp. 2353--2367, 2024.

\bibitem{37}
M.~Fan, W.~Deng, Y.~Zhu, and C.~Zhang, ``Applications and advances of pet imaging in alzheimer's disease,'' \emph{Int J Radiat Med Nucl Med}, vol.~42, no.~6, pp. 553--558, 2018.

\bibitem{38}
S.~Chen, Z.~Cao, A.~Nandi, N.~Counts, L.~Jiao, K.~Prettner, M.~Kuhn, B.~Seligman, D.~Tortorice, D.~Vigo \emph{et~al.}, ``The global macroeconomic burden of alzheimer's disease and other dementias: estimates and projections for 152 countries or territories,'' \emph{The Lancet Global Health}, vol.~12, no.~9, pp. e1534--e1543, 2024.

\bibitem{39}
C.~R. Jack~Jr, J.~S. Andrews, T.~G. Beach, T.~Buracchio, B.~Dunn, A.~Graf, O.~Hansson, C.~Ho, W.~Jagust, E.~McDade \emph{et~al.}, ``Revised criteria for diagnosis and staging of alzheimer's disease: Alzheimer's association workgroup,'' \emph{Alzheimer's \& Dementia}, vol.~20, no.~8, pp. 5143--5169, 2024.

\bibitem{40}
A.~Association \emph{et~al.}, ``2018 alzheimer's disease facts and figures,'' \emph{Alzheimer's \& Dementia}, vol.~14, no.~3, pp. 367--429, 2018.

\bibitem{41}
T.~O. Frizzell, M.~Glashutter, C.~C. Liu, A.~Zeng, D.~Pan, S.~G. Hajra, R.~C. D’Arcy, and X.~Song, ``Artificial intelligence in brain mri analysis of alzheimer’s disease over the past 12 years: A systematic review,'' \emph{Ageing Research Reviews}, vol.~77, p. 101614, 2022.

\bibitem{42}
M.~Woodward, D.~A. Bennett, T.~Rundek, G.~Perry, and T.~Rudka, ``The relationship between hippocampal changes in healthy aging and alzheimer’s disease: a systematic literature review,'' \emph{Frontiers in Aging Neuroscience}, vol.~16, p. 1390574, 2024.

\bibitem{43}
Z.~Fang, S.~Zhu, Y.~Chen, B.~Zou, F.~Jia, L.~Qiu, C.~Liu, Y.~Huang, X.~Feng, F.~Qin \emph{et~al.}, ``Gfe-mamba: Mamba-based ad multi-modal progression assessment via generative feature extraction from mci,'' \emph{arXiv preprint arXiv:2407.15719}, 2024.

\bibitem{44}
S.~Zhu, Y.~Chen, S.~Jiang, W.~Chen, C.~Liu, Y.~Wang, X.~Chen, Y.~Ke, F.~Qin, C.~Wang \emph{et~al.}, ``Xlstm-hved: Cross-modal brain tumor segmentation and mri reconstruction method using vision xlstm and heteromodal variational encoder-decoder,'' in \emph{2025 IEEE 22nd International Symposium on Biomedical Imaging (ISBI)}.\hskip 1em plus 0.5em minus 0.4em\relax IEEE, 2025, pp. 1--5.

\bibitem{zhang2024tc}
C.~Zhang, Y.~Chen, Z.~Fan, Y.~Huang, W.~Weng, R.~Ge, D.~Zeng, and C.~Wang, ``Tc-diffrecon: Texture coordination mri reconstruction method based on diffusion model and modified mf-unet method,'' in \emph{2024 IEEE International Symposium on Biomedical Imaging (ISBI)}.\hskip 1em plus 0.5em minus 0.4em\relax IEEE, 2024, pp. 1--5.

\bibitem{ge2024tc}
R.~Ge, X.~Yu, Y.~Chen, G.~Zhou, F.~Jia, S.~Zhu, J.~Jia, C.~Zhang, Y.~Sun, D.~Zeng \emph{et~al.}, ``Tc-kanrecon: High-quality and accelerated mri reconstruction via adaptive kan mechanisms and intelligent feature scaling,'' \emph{arXiv preprint arXiv:2408.05705}, 2024.

\bibitem{chen2024scunet++}
Y.~Chen, B.~Zou, Z.~Guo, Y.~Huang, Y.~Huang, F.~Qin, Q.~Li, and C.~Wang, ``Scunet++: Swin-unet and cnn bottleneck hybrid architecture with multi-fusion dense skip connection for pulmonary embolism ct image segmentation,'' in \emph{Proceedings of the IEEE/CVF Winter Conference on Applications of Computer Vision}, 2024, pp. 7759--7767.

\bibitem{liu2024spiking}
M.~Liu, J.~Gan, R.~Wen, T.~Li, Y.~Chen, and H.~Chen, ``Spiking-diffusion: Vector quantized discrete diffusion model with spiking neural networks,'' in \emph{2024 5th International Conference on Computer, Big Data and Artificial Intelligence (ICCBD+ AI)}.\hskip 1em plus 0.5em minus 0.4em\relax IEEE, 2024, pp. 627--631.

\bibitem{zhu2025towards}
Z.~Zhu, S.~Jiang, J.~Zheng, Y.~Li, Y.~Chen, M.~Zhao, W.~Gu, F.~Qin, J.~Wang, and G.~Yu, ``Towards accurate and interpretable neuroblastoma diagnosis via contrastive multi-scale pathological image analysis,'' \emph{arXiv preprint arXiv:2504.13754}, 2025.

\bibitem{chen2022neuroimaging}
J.~Chen, S.~Wang, R.~Chen, and Y.~Liu, ``Neuroimaging biomarkers and cognition in alzheimer's disease spectrum,'' p. 848719, 2022.

\bibitem{kang20233dnest}
X.~Kang and Y.~Liu, ``3dnest: A hierarchical local self-attention model for alzheimer’s disease diagnosis,'' in \emph{2023 International Conference on Neuromorphic Computing (ICNC)}.\hskip 1em plus 0.5em minus 0.4em\relax IEEE, 2023, pp. 150--158.

\bibitem{oh2025cerebrospinal}
H.~S.-H. Oh, D.~Y. Urey, L.~Karlsson, Z.~Zhu, Y.~Shen, A.~Farinas, J.~Timsina, M.~R. Duggan, J.~Chen, I.~H. Guldner \emph{et~al.}, ``A cerebrospinal fluid synaptic protein biomarker for prediction of cognitive resilience versus decline in alzheimer’s disease,'' \emph{Nature medicine}, pp. 1--12, 2025.

\bibitem{weiner2015impact}
M.~W. Weiner, D.~P. Veitch, P.~S. Aisen, L.~A. Beckett, N.~J. Cairns, J.~Cedarbaum, M.~C. Donohue, R.~C. Green, D.~Harvey, C.~R. Jack~Jr \emph{et~al.}, ``Impact of the alzheimer's disease neuroimaging initiative, 2004 to 2014,'' \emph{Alzheimer's \& Dementia}, vol.~11, no.~7, pp. 865--884, 2015.

\bibitem{fowler2021fifteen}
C.~Fowler, S.~R. Rainey-Smith, S.~Bird, J.~Bomke, P.~Bourgeat, B.~M. Brown, S.~C. Burnham, A.~I. Bush, C.~Chadunow, S.~Collins \emph{et~al.}, ``Fifteen years of the australian imaging, biomarkers and lifestyle (aibl) study: progress and observations from 2,359 older adults spanning the spectrum from cognitive normality to alzheimer’s disease,'' \emph{Journal of Alzheimer's disease reports}, vol.~5, no.~1, pp. 443--468, 2021.

\bibitem{song2024learning}
T.~Song, G.~Jin, P.~Li, K.~Jiang, X.~Chen, and J.~Jin, ``Learning a spiking neural network for efficient image deraining,'' \emph{arXiv preprint arXiv:2405.06277}, 2024.

\end{thebibliography}


\end{document}